%% file: main.tex
\documentclass[review,hidelinks,onefignum,onetabnum]{siamart220329}
\nolinenumbers

\usepackage{subfigure}
\usepackage{caption}
\usepackage{multirow}
\usepackage{mathrsfs}
\usepackage{enumerate}
\usepackage{graphicx,epsfig}
\usepackage{epstopdf}
\usepackage{booktabs} 
\usepackage{siunitx}
\usepackage{amssymb}

\usepackage{booktabs}
\usepackage{siunitx}
\newtheorem{example}[theorem]{Example}

\newcommand{\bx}{\mathbf{x}}
\newcommand{\by}{{\mathbf y}}

\input{ex_shared}

\ifpdf
\hypersetup{
  pdftitle={Unsupervised operator learning approach for dissipative equations via Onsager principle},
  pdfauthor={Zhipeng Chang, Zhenye Wen, Xiaofei Zhao}
}
\fi

\begin{document}
\begin{sloppypar}

\maketitle
\begin{abstract}
Existing operator learning methods rely on supervised training with high-fidelity simulation data, introducing significant computational cost. In this work, we propose the deep Onsager operator learning (DOOL) method, a novel unsupervised framework for solving dissipative equations. Rooted in the Onsager variational principle (OVP), DOOL trains a deep operator network by directly minimizing the OVP-defined Rayleighian functional, requiring no labeled data, and then proceeds in time explicitly through conservation/change laws for the solution.
Another key innovation here lies in the spatiotemporal decoupling strategy: the operator's trunk network processes spatial coordinates exclusively, thereby enhancing training efficiency, while integrated external time stepping enables temporal extrapolation.
Numerical experiments on typical dissipative equations validate the effectiveness of the DOOL method, and systematic comparisons with supervised DeepONet and MIONet demonstrate its enhanced performance.
Extensions are made to cover the second-order wave models with dissipation that do not directly follow OVP. 
\end{abstract}



\section{Introduction}
In reality, dissipation induced by friction and/or diffusion occurs widely.
Dissipative systems, e.g., gradient flows and damped wave models, typically characterized by irreversible energy decay, govern a wide range of important physical phenomena in the fields of science and engineering \cite{Doi,Kondo, Petrov}.  
For efficient and accurate simulations, classical numerical methods have been extensively developed to solve dissipative PDE models, particularly gradient flows \cite{Hochbruck,Shen2018,Shen2012,Yang2017}. 
In recent years, the rapid development of artificial intelligence has given rise to a variety of PDE solvers based on deep learning, e.g., the physics-informed neural networks \cite{pinns}, the deep Ritz method \cite{DeepR}, the deep Galerkin method \cite{DGM}, the weak adversarial network \cite{WAN} and the random feature method \cite{RFM}. These methods by neural networks, approximate solutions of PDE under fixed configuration. 
Although successful, they lack generalization ability with respect to new model configurations, such as the initial value and the physical  parameters. Retraining for each new configuration incurs prohibitive computational costs. 

This motives the operator learning that establishes the mapping from the input function space (e.g., initial conditions, source terms) to the output solution space. 
Guided by the universal operator approximation theorem \cite{Chen1995}, numerous operator learning methods have been developed, such as the Fourier neural operator (FNO) \cite{FNO}, the deep operator
network (DeepONet) \cite{DeepONet}, the graph kernel network \cite{GKN}, and the random feature model \cite{RFMO}. 
The FNO introduced in \cite{FNO} leverages Fourier transforms to perform global convolutions in spectral space, enabling efficient PDE solution operator learning on structured grids. It is particularly suitable for regular domains and periodic problems, and has subsequently been extended to Geo-FNO \cite{GeoFNO} for arbitrary geometries, OPNO \cite{OPNO} for non-periodic boundary conditions, as well as various other improvements from different perspectives \cite{GT,spectral,MgNO,PINO}. The other major approach DeepONet designs a branch-trunk architecture to learn functional and spatiotemporal information separately \cite{DeepONet}. This network gives it flexibility to handle complex and unstructured input, applying to many scientific computing problems \cite{DeepMMnet, NF,Li2024water,GE2021}. 
Subsequent developments include the MIONet \cite{Jin} for multiple inputs, the Bayesian DeepONet \cite{BayesDeepONet} for stochastic problems, and many other refinements, such as \cite{PODDeepONet,piDeepONet}.

Despite significant advancements in operator learning, all existing frameworks are based on supervised learning, which fundamentally relies on high-fidelity labeled data. Such labeled data are typically generated by running classical numerical solvers for the underlying PDE models---a process that is often computationally expensive and labor-intensive. 
This motivates us to develop an unsupervised operator learning approach, and we shall focus on dissipative PDEs in this work. 
The key technique we employ is the Onsager variational principle (OVP), originally proposed by Onsager in his seminal work on the dynamics of irreversible processes \cite{Onsager1}. It belongs to the broader family of least-action principles. This powerful principle has been successfully applied both in modeling \cite{Doi,Doi2015} and in numerical simulations \cite{Onsagerchen}. More recently, OVP has been integrated with deep learning to infer dynamics from data \cite{VONN,OnsagerNet}, 
and its generalization, the Onsager-Machlup principle has been implemented to solve PDEs \cite{OM2025, NNaction2022}.

In this work, based on OVP, we will propose a fully unsupervised operator learning approach for solving dissipative equations, named the deep Onsager operator learning (DOOL) method. 
It trains a deep operator network, which in principle can be in the user's favor but is fixed as the branch-trunk architecture in this work, by minimizing the Rayleighian functional under OVP to learn the mapping from an input state field to its flux. The trained operator network is then used to update the state field in time by externally discretizing a conservation/change law.
Such an approach possesses the following appealing advantages which are validated through our numerical experiments:
\begin{itemize}
    \item[1)] \textbf{Unsupervised training}: The loss function is just the Rayleighian functional, which is clearly defined under OVP for a dissipative system. There is no need for any labeled data.

   \item[2)]  \textbf{Training efficiency}: In the operator network used to learn constitutive relationship, the trunk network processes only the spatial coordinates, rather than unified spatiotemporal inputs. This reduces the number of sampling points and the dimensionality of the input, thereby improving training efficiency.

      \item[3)] \textbf{Temporal extrapolation ability}: The temporal update is performed through an external time stepping that is not restricted to a prescribed time interval for training. The spatiotemporal decoupling framework supports naturally the extrapolation to arbitrary time, as long as the operator network is trained in the functional space in which the dissipative model is well-posed. 
\end{itemize}

The numerical experiments for validation include the heat equation, the Fokker-Planck equation, the Cahn-Hilliard equation, and the Allen-Cahn equation. 
Comparisons in accuracy and efficiency are made with the supervised DeepONet and MIONet. 
It is also noteworthy that monotonic energy dissipation has been numerically observed and analyzed.
As an application, DOOL can be used for parameter inversion problems, as demonstrated on the Cahn-Hilliard equation. In the end, the extension is made to cover the second-order wave models with dissipation that do not directly follow OVP. 

The rest of this paper is structured as follows. Section \ref{sec:pre} provides some necessary preliminaries on the OVP and the operator network. Section \ref{sec:DOOL} details the implementation of DOOL and presents its performance. Section \ref{sec:compare} evaluates the generalization capability of DOOL and conducts systematic comparisons. Applications and extensions are given in Section \ref{sec:extension}, and some conclusions are drawn in Section \ref{sec:conclusions}.

\section{Preliminary}\label{sec:pre}
In this section, we briefly introduce some preliminary knowledge that is fundamental for proposing our method. 
The first is a general principle to express diffusive systems, and the second is a network for operator learning. 

\subsection{Onsager variational principle (OVP)}
Characterizing a physical system in terms of a comprehensive set of variables $\boldsymbol{u} = (u_1,\cdots,u_s)^\top$, assuming that the system undergoes an irreversible process within a linear response mechanism and neglecting inertial effects, one can derive the process using the OVP \cite{Onsagerchen,Doi}. 
For systems with conserved and nonconserved parameter, the OVP is described as follows:

\begin{enumerate}[(i)]
	
	\item \emph{System for conserved parameters:} In  many physical systems, certain variables (e.g., the mass) are conserved in time. They are referred as the systems for conserved parameter, and the conservation laws can be written as
	\begin{equation}\label{eq:conservation_eq}
		\partial_t u_k + \nabla \cdot  \boldsymbol{j}_k = 0,\quad k = 1,\cdots,s,
	\end{equation}
	where $\boldsymbol{j}_k$ denotes the flux corresponding to $u_k$. The dynamics here can be defined if $\boldsymbol{j}:=(\boldsymbol{j}_1^\top, \cdots, \boldsymbol{j}_s^\top)^\top$ is determined (referred as the constitutive relation). The OVP then states that this is determined by minimizing the Rayleighian functional with respect to $\boldsymbol{j}$ under the constraints of conservation equations:
    \begin{equation*}
     \boldsymbol{j} \in \arg\min \big\{ \mathcal{R}(\boldsymbol{u}, \partial_t \boldsymbol{u}, \boldsymbol{j}) \ : \ \partial_t \boldsymbol{u} + \nabla \cdot \boldsymbol{j} = \mathbf{0} \big\},   
    \end{equation*}
    where the Rayleighian functional $\mathcal{R}$ is given by
	\begin{equation}\label{eq:Rayleighian conserved}
		\mathcal{R}(\boldsymbol{u}, \partial_t \boldsymbol{u}, \boldsymbol{j}) :=   \dot{\mathcal{E}} (\boldsymbol{u}, \partial_t \boldsymbol{u}) + \Phi(\boldsymbol{u}, \boldsymbol{j}),
	\end{equation}
	with $\mathcal{E}(\boldsymbol{u})$ the free energy functional, $\dot{\mathcal{E}}$ the time derivative of $\mathcal{E}$ and $\Phi$ the energy dissipation function. Typical dissipative equations within such a framework include the heat equation, Fokker-Planck equation and Cahn-Hilliard equation. 
	
	\item \emph{System for nonconserved parameters:} For the system with nonconserved parameter, where no conservation laws hold, the Rayleighian functional is defined as 
	\begin{equation*}
		\mathcal{R}(\boldsymbol{u}, \partial_t \boldsymbol{u}) :=   \dot{\mathcal{E}} (\boldsymbol{u}, \partial_t \boldsymbol{u}) + \Phi(\boldsymbol{u}, \partial_t \boldsymbol{u}).
	\end{equation*}
	The OVP is this case gives the dynamics of the system by minimizing $\mathcal{R}$ with respect to $\partial_t \boldsymbol{u}$:
    \begin{equation}\label{ovp noncon}
    \partial_t\boldsymbol{u} \in \arg\min \big\{ \mathcal{R}(\boldsymbol{u}, \partial_t \boldsymbol{u}) \big\},
   \end{equation}
    which infers the change rate of the state variable. 
    The typical example of such system is the Allen-Cahn equation.
	
\end{enumerate}

Utilizing the intrinsic minimization in the OVP, we shall design an unsupervised operator learning approach for solving general dissipative systems. 

\subsection{Deep operator network}

The network that we shall adopt to approximate nonlinear operators between infinite-dimensional function spaces is
the architecture of DeepONet \cite{DeepONet}. It is defined as follows. 
Consider an operator $\mathcal{G}: \mathcal{W} \to \mathcal{V}$ that maps an input function $w \in \mathcal{W}$ to an output function $v \in \mathcal{V}$, where $\mathcal{W}$ and $\mathcal{V}$ are appropriate function spaces defined on domains $\Omega_w \subset \mathbb{R}^{d_w}$ and $\Omega_v \subset \mathbb{R}^{d_v}$, respectively, for  $d_w,d_v \in \mathbb{N}_{+}$.
The DeepONet approximates $\mathcal{G}(w)$ as: for any $\mathbf{y} \in \Omega_v$, 
\begin{equation}
	\label{DeepONet}
	\mathcal{G}(w)(\mathbf{y}) \approx \mathcal{G}_{\theta}(w, \mathbf{y}) = \sum_{k=1}^{p} \underbrace{\mathfrak{b}_k \left( w(\mathbf{s}_1), w(\mathbf{s}_2), \dots, w(\mathbf{s}_m);{\theta_{\mathfrak{b}}} \right)}_{\text{branch network}}\quad \cdot \underbrace{\mathfrak{t}_k (\mathbf{y};{\theta_{\mathfrak{t}}})}_{\text{trunk network}},
\end{equation}
where $p\in\mathbb{N}_{+}$ {represents the output width of the branch and trunk network}, $\{\mathbf{s}_l\}_{l=1}^m \subset \Omega_w$ {are sensors}, and $\theta\ {=\{\theta_{\mathfrak{b}},\,\theta_{\mathfrak{t}}\}}$ denotes the trainable parameters in the network. 

The {branch network} encodes the input function $w$ sampled at fixed sensors $\{\mathbf{s}_l\}_{l=1}^m $. It takes  $(w(\mathbf{s}_1), w(\mathbf{s}_2), \dots, w(\mathbf{s}_m))^\top$ as input and outputs a $p$-dimensional latent representation $\mathfrak{b} = (\mathfrak{b}_1, \mathfrak{b}_2, \dots, \mathfrak{b}_p)\in \mathbb{R}^p$, typically implemented as a multilayer perceptron (MLP) \cite{Rosenblatt}.
The {trunk network} takes spatial/temporal coordinates $\mathbf{y} \in \Omega_v$ as input and outputs the vector $\mathfrak{t} = (\mathfrak{t}_1, \mathfrak{t}_2, \dots, \mathfrak{t}_p) \in \mathbb{R}^p$ which is also an MLP adaptive to the output domain.
 The prediction of the final operator is formed by the inner product of the branch and trunk outputs, providing a continuous representation in $\mathbf{y}$. 
For simplicity, the same number of neurons is used in each hidden layer of both the branch and trunk networks, with identical depth $L$ and width $W$.

DeepONet has been widely applied ever since its birth to solve various types of differential equations \cite{DeepMMnet,NF,Li2024water, GE2021}, while the training of DeepONet in existing studies is mainly based on supervised learning. That is, the operator $\mathcal{G}_\theta$ is learned by minimizing the difference between the prediction and the data labeled from the true solution. 
Let us take a time-space-dependent problem as an example with $\mathbf{u}=\mathcal{G}(w)(\mathbf{x},t)$ the true solution for illustration.
A typical case occurs when $w$ represents the initial condition and $(w(\mathbf{s}_1), w(\mathbf{s}_2), \dots, w(\mathbf{s}_m))^\top$ denotes its discrete representation in spatial coordinates.
In addition to the sampled spatial/temporal grid points $\{(\mathbf{x}_k,t_n)\}_{k,n=1}^{N_x,N_t}\subset \Omega_v$, a number of characteristic functions $\{w^{(b)}\}_{b=1}^{N_b}$ need to be sampled from $\mathcal{W}$, with $N_x,N_t,N_b\in\mathbb{N}_+$. All of these together form a dataset: $\{(w^{(b)}, \{(\mathbf{x}_k, t_n)\}_{k,n=1}^{N_x,N_t}), \{\mathcal{G}(w^{(b)})(\mathbf{x}_k, t_n)\}_{k,n=1}^{N_x,N_t} \}_{b=1}^{N_b}$, and the parameters $\theta$ of DeepONet are optimized by minimizing
\begin{equation}\label{eq:loss_deeponet}
\text{Loss}(\theta) = \frac{1}{N_b\times N_x \times N_t} \sum_{b=1}^{N_b} \sum_{k=1}^{N_x} \sum_{n=1}^{N_t} \left\| \mathcal{G}(w^{(b)})(\mathbf{x}_k, t_n) -\mathcal{G}_\theta(w^{(b)})(\mathbf{x}_k, t_n) \right\|_2^2.
\end{equation}
This supervised approach, while effective, relies on the availability of high-fidelity labeled data that may be costly or infeasible to obtain. 

\begin{figure}[h!]
\centering
    \psfig{figure=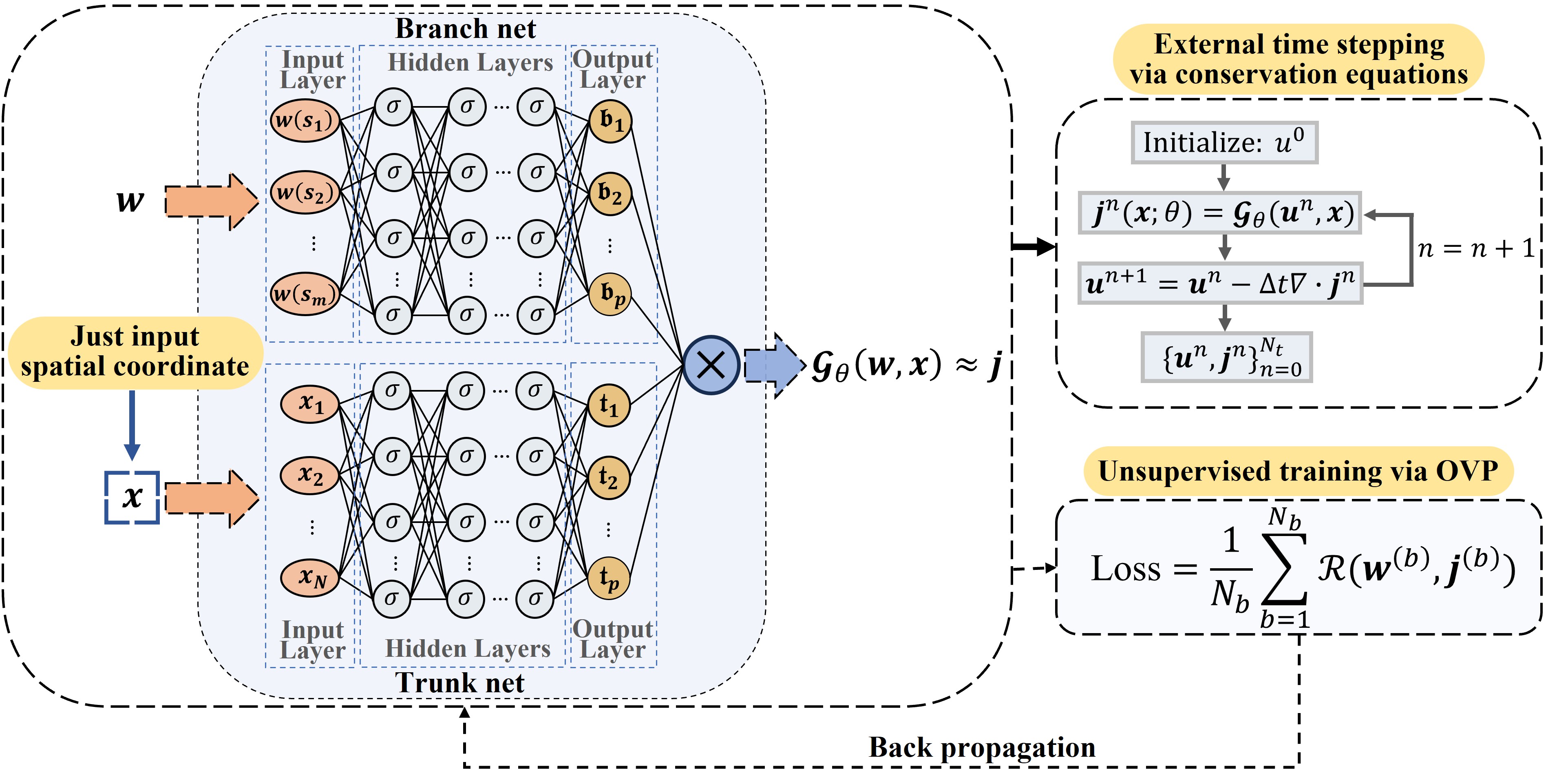,width=5in,height=2.3in,angle=0}
		\hspace{-0.1in}

	\caption{Illustration of the architecture of DOOL.}\label{fig:arc}
\end{figure}

In this study, we aim to develop an unsupervised training approach for DeepONet based on the OVP that eliminates the need for labeled data. It employs the basic network structure of DeepONet to approximate the time-independent mapping from a state ${w}$ to the flux  $\boldsymbol{j}$,  thereby establishing a \textbf{deep Onsager operator learning (DOOL)} method for solving dissipative differential equations.
Moreover, DOOL can improve upon the basic DeepONet structure through: \begin{itemize}
    \item The trunk and branch network input only spatial domain information, rather than unified spatiotemporal processing, which reduces the number of sampled points and input; 
\item After training, the temporal information is updated through an explicit time stepping, which is not restricted to interpolation within a prescribed time interval.
\end{itemize}  
The architecture of the DOOL method for the system for conserved parameters is demonstrated in Fig. \ref{fig:arc}, and the detailed development is presented in the next section. It should be noted that other operator networks, such as the FNO \cite{FNO}, are also feasible for DOOL, which will be addressed in other works. 

\section{Deep Onsager operator learning (DOOL) method}\label{sec:DOOL}
In this section, we present the detailed framework of our DOOL method. Here we shall focus on systems for conserved parameters for illustrations: 
\begin{equation}\label{eq dynamics}
\partial_t \boldsymbol{u}(\mathbf{x},t) + \nabla \cdot \boldsymbol{j}(\mathbf{x},t) = \mathbf{0},\quad \mathbf{x}\in\Omega,\ t>0,
\end{equation}
where boundary conditions (such as periodic or homogeneous Neumann or Dirichlet) 
that do not contribute when integrating by parts are imposed at $\partial \Omega$. The case of systems for non-conserved parameter will be addressed later in Section \ref{subsec:noncon}.

\subsection{Unsupervised operator learning through Rayleighian}
\label{subsubsec:unsupervised_training}

We are concerned with the relationship between the state field $\boldsymbol{u}$ and the corresponding flux $\boldsymbol{j}$ at any given time instant---represented by the operator $\mathcal{G}: \boldsymbol{u}(\bx,t) \to \boldsymbol{j}(\bx,t)$. 
Based on the OVP, this operator can be read as minimizing the Rayleighian functional $\mathcal{R}$ \eqref{eq:Rayleighian conserved} under a given $\mathbf{u}$ and the conservation constraint \eqref{eq:conservation_eq}:
\begin{align}
	\boldsymbol{j} &\in \arg\min \big\{ \mathcal{R}(\boldsymbol{u}, \partial_t \boldsymbol{u}, \boldsymbol{j}) \ : \ \partial_t \boldsymbol{u} + \nabla \cdot \boldsymbol{j} = \mathbf{0} \big\} \notag\\
	&= \arg\min \big\{ \mathcal{R}(\boldsymbol{u}, -\nabla \cdot \boldsymbol{j}, \boldsymbol{j}) \big\}  \notag\\
	&\triangleq \arg\min \big\{ \widetilde{\mathcal{R}}(\boldsymbol{u}, \boldsymbol{j}) \big\} \label{eq:reduced_form}.
\end{align}
The operator network $\mathcal{G}_{\theta}$ (\ref{DeepONet}) is set to approximate $\mathcal{G}$, i.e., $\mathcal{G}_{\theta}(\boldsymbol{u}, \cdot) \approx \mathcal{G}(\boldsymbol{u}) =  \boldsymbol{j}$. 
Plugging it into $\widetilde{\mathcal{R}}$ in  \eqref{eq:reduced_form},
the optimized $\theta$ can be obtained through the natural minimization of the Rayleighian function. 
This thus forms an unsupervised manner for training  $\mathcal{G}_{\theta}$ via the OVP.
The training details are explained below. 

The trunk network of (\ref{DeepONet}) takes spatial grid points $\mathbf{x}\in\Omega$ as input, where uniform grids will be used in later experiments for convenience. The branch network in (\ref{DeepONet}) receives information about the state field $\boldsymbol{u}$ through its coefficient vector representation under a chosen set of basis functions in the concerned function space. Suppose that the dynamics of (\ref{eq dynamics}) keeps $\boldsymbol{u}$ in a certain functional space $\mathcal{W}$. With $\left\{\psi_{\boldsymbol{k}}(\mathbf{x})\right\}$ denoting a basis of $\mathcal{W}$, 
\begin{equation}\label{eq:basis}
  \boldsymbol{u}\approx \sum_{\boldsymbol{k}\in \mathbb{Z}^d,\,|\boldsymbol{k}|\leq K}c_{\boldsymbol{k}}\psi_{\boldsymbol{k}},\quad K\in\mathbb{N}_+,  
\end{equation}
where convergence happens as $K\to\infty$, and  $c_{\boldsymbol{k}} \in \mathbb{C}$ is the coefficient to practically input to the branch network. In this paper, we will implement two sets of basis. 
{1) Fourier basis:} Consider  $\Omega = [-I,I]^d \subset \mathbb{R}^d$ with periodic boundary conditions and $\psi_{\boldsymbol{k}}(\mathbf{x})=\exp\{i\pi \mathbf{k}\cdot\mathbf{x}/I\}$ for $\boldsymbol{u} \in L^2(\Omega)$. {2) Hermite basis:} Consider $d=1$ for simplicity with zero boundary conditions and 
the normalized Hermite basis function 
$
\psi_k(x) = H_k(x) \cdot \mathrm{e}^{-x^2/2}/\sqrt{2^k k! \sqrt{\pi}},
$
where $H_k(x) = (-1)^k\mathrm{e}^{x^2}\frac{d^k}{dx^k}\mathrm{e}^{-x^2}$ is the $k$-th Hermite polynomial with $k \in \mathbb{N}$.

To enable $\mathcal{G}_{\theta}$ capturing general information of $\mathcal{W}$, we sample each $c_{\boldsymbol{k}}$ in (\ref{eq:basis}) on some rectangle of complex plane randomly according to the uniform distribution. Note that the smoothness of $\boldsymbol{u}$ (especially for diffusive models) indicates the (fast) decay of $c_{\boldsymbol{k}}$ as $|\boldsymbol{k}|$ increases. Thus, 
a helpful technique in the sampling process involves assigning wider rectangles to low-frequency components (capturing large-scale features), while constraining high-frequency components (representing small-scale features) to narrower rectangles. This helps in efficiently generating the dataset with sufficient spectral characteristics. Suppose that a set of $N_b\in \mathbb{N}_+$ samples is generated, and then one gets $ \displaystyle \boldsymbol{u}^{(b)}=\sum_{\boldsymbol{k}\in \mathbb{Z}^d,\,|\boldsymbol{k}|\leq K}c^{(b)}_{\boldsymbol{k}}\psi_{\boldsymbol{k}}$, for $b=1,\ldots,N_b$. 

The sampled $\{ \boldsymbol{u}^{(b)} \}_{b=1}^{N_b}$ is used to calculate the loss function, given as follows.
The approximate fluxes are first calculated by forward propagation: $\boldsymbol{j}^{(b)}(\mathbf{x}{;\theta}) = \mathcal{G}_\theta \big( \boldsymbol{u}^{(b)}, \mathbf{x} \big)$ for each sample $b=1,\dots,N_b$, and then we can obtain
the corresponding Rayleighian function $\widetilde{\mathcal{R}}^{(b)} = \widetilde{\mathcal{R}}\big( \boldsymbol{u}^{(b)}, \boldsymbol{j}^{(b)} \big)$ defined by \eqref{eq:reduced_form} for each state-flux pair. As stated by the OVP, the minimization of  $\widetilde{\mathcal{R}}^{(b)}$ should be done for each $b$. This motivates the following definition of the loss function:
\begin{equation}\label{loss_r_sum}
	\text{Loss}_{{\mathcal{R}}}(\theta) = \frac{1}{N_b} \sum_{b=1}^{N_b} \widetilde{\mathcal{R}}^{(b)}.
\end{equation}
Consequently, the operator learning is formulated as the following unconstrained optimization problem:
\begin{equation}\label{min_theta}
	\min_\theta\left\{ \text{Loss}_{{\mathcal{R}}}(\theta)  \right\}.
\end{equation}
To solve \eqref{min_theta}, 
we shall adopt the Adam optimizer \cite{adamKingma2015} with the Xavier method \cite{glorot2010} for parameter initialization in our numerical experiments.
The learning rate is set as $\tau = 5\times 10^{-4}$ in the studies, unless otherwise specified. 

The direct minimization of the Rayleighian functional from \eqref{eq:reduced_form} avoids the need of prior flux data as labels for supervision, where such labeled data usually have to be prepared by repeatably running some standard numerical solver towards the targeted PDEs. This brings significant convenience.


\subsection{PDE solver}
The trained operator network $\mathcal{G}_\theta$ maps an input state $\boldsymbol{u}$ to the corresponding flux $\boldsymbol{j}$. Based on it, by temporally discretizing the conservation equation \eqref{eq dynamics}, we can numerically evolve the system starting from any given initial state, predicting the state field $\boldsymbol{u}(\bx,t)$ and the flux $\boldsymbol{j}(\bx,t)$ up to an arbitrary time.

Let $t_n = n\Delta t$ for $n \geq 0$ with $\Delta t > 0$ the time step size, and denote $\boldsymbol{u}^n(\mathbf{x})\approx \boldsymbol{u}(\mathbf{x},t_n),\boldsymbol{j}^n(\mathbf{x})\approx \boldsymbol{j}(\mathbf{x},t_n)$. Given an initial condition $\boldsymbol{u}_0=\boldsymbol{u}^0$ at $t = 0$, the flux field for each $n>0$ is predicted using the trained operator network:
\begin{equation}\label{dool jn}
	\boldsymbol{j}^{n}(\mathbf{x}) = \mathcal{G}_\theta(\boldsymbol{u}^{n}, \mathbf{x}).
\end{equation}
The conservation equation \eqref{eq:conservation_eq} can be discretized in time simply by the forward Euler scheme:
\begin{equation}\label{eq:time_d}
	\frac{\boldsymbol{u}^{n+1} - \boldsymbol{u}^{n}}{\Delta t} + \nabla \cdot \boldsymbol{j}^{n} = \mathbf{0},
\end{equation}
which provides the state update formula $\boldsymbol{u}^{n+1} = \boldsymbol{u}^{n} - \Delta t \, \nabla \cdot \boldsymbol{j}^{n}$ for $n=0,1,\ldots$. Starting from $\boldsymbol{u}^{0}$, this iterative procedure computes the solution fields $\boldsymbol{u}$ and $\boldsymbol{j}$ at all subsequent time steps, which completes the DOOL method.

The forward Euler scheme is applied here because it offers an explicit and simple approximation that is accurate enough. As a matter of fact, the dominant error of DOOL comes from the approximation error of operator learning based on our numerical experience, and the temporal discretization error from (\ref{eq:time_d}) is comparatively negligible. 
More schemes, including implicit methods and higher-order methods, may be considered in the future. 
The spatial derivatives involved for problems with periodic boundary conditions can be easily computed using the fast Fourier transform. Alternatively, they can be computed by network backpropagation. Our numerical experience confirms that both approaches are feasible.

The DOOL method for solving dissipative PDEs is summarized in Algorithm \ref{alg:workflow}. 
It exhibits three key features: \begin{itemize}
 \item Once the operator network $\mathcal{G}_\theta$ (\ref{DeepONet}) is trained, it achieves fast computational speed in predicting the flux through the forward propagation (\ref{dool jn});  
 
 \item The time stepping (\ref{eq:time_d}) is not restricted to a fixed time interval, and so the extrapolation in time is natural;
 
 \item The operator network $\mathcal{G}_\theta$ is trained in a chosen functional space for $\boldsymbol{u}$ and \eqref{eq:time_d} can start from any initial value from the concerned space, which means that the generalization with respect to the initial value of (\ref{eq dynamics}) is naturally available. 

\end{itemize}

\begin{algorithm}[h!]
	\renewcommand{\algorithmicrequire}{\textbf{Input:}}
	\renewcommand{\algorithmicensure}{\textbf{Output:}}
	\caption{DOOL for solving dissipative PDEs}
	\label{alg:workflow}
    Parameters: Number of function samples $N_b$; Frequency truncation order $K$; Time step $\Delta t$
	\begin{algorithmic}[1]
		\REQUIRE Initial state $\boldsymbol{u}^{0}$; Space point $\bx$ for trunk 
		\ENSURE Operator network $\mathcal{G}_\theta : \boldsymbol{u} \mapsto \boldsymbol{j}$; Solution fields $\{ \boldsymbol{u}^{n}, \boldsymbol{j}^{n} \} _{n\geq0}$
		
		 \textbf{Part I: Unsupervised operator network training}
		\STATE \emph{Unlabeled training samples generation:} generate coefficient vectors $\{\mathbf{c}^{(b)}\}_{b=1}^{N_b}$ and compute corresponding 
$\{\boldsymbol{u}^{(b)}\}_{b=1}^{N_b}$
		\STATE \emph{Train operator network through Rayleighian:} initialize $\theta$
		\FOR{$epoch = 1$ \textbf{to} $epoch = N_e$}
		\STATE for each $b$, forward pass $\boldsymbol{j}^{(b)}=\mathcal{G}_\theta(\mathbf{u}^{(b)},\bx)$
		and compute Rayleighian $\widetilde{\mathcal{R}}^{(b)} = \widetilde{\mathcal{R}} (\boldsymbol{u}^{(b)},\boldsymbol{j}^{(b)})$
		\STATE compute the loss function $\text{Loss}_{\mathcal{R}}(\theta)$ by \eqref{loss_r_sum}
		\STATE update $\theta$ via the optimizer
		\ENDFOR
		
		 \textbf{Part II: Time stepping by conservation equations}
		\STATE \textbf{Initialize:} $\boldsymbol{u}^{0} \gets \boldsymbol{u}_0$, 
        $t \gets t_0$
		\FOR{$n \geq 0$ }
		\STATE predict flux $\boldsymbol{j}^{n} = \mathcal{G}_\theta(\boldsymbol{u}^{n}, \mathbf{x})$
		\STATE update solution $\boldsymbol{u}^{n+1} = \boldsymbol{u}^{n} - \Delta t \, \nabla \cdot \boldsymbol{j}^{n}$
		\STATE advance time $t \gets t + \Delta t$
		\ENDFOR
	\end{algorithmic}
\end{algorithm}

\subsection{Numerical experiments}\label{sec:numerical experiments} 
We now illustrate the performance of the proposed DOOL in Algorithm \ref{alg:workflow}, using a series of examples\footnote{Codes are available at {https://github.com/wzhy0777/DOOL}, with computations performed in PyTorch sequentially on a laptop with an Intel Core i9-14900HX processor and 32 GB of memory.} including the pure diffusion process and mixed diffusion dynamics with convection or reaction.

\begin{example}[Heat equation]\label{ex:Linear-H}
	Let us begin with the simplest pure diffusion case, a standard one-dimensional (1D) heat equation:
	\begin{equation}\label{eq:diffusion}
		\begin{cases}
			\partial_tu(x,t) \;=\; \partial_{xx}u(x,t), & x\in(-I, I),\; t>0,\\
			u(L,t)=u(-L,t)=C_0,     & t\geq0,\\
			u(x,0)=u_0(x),       & x\in[-I, I].
		\end{cases}
	\end{equation}
		Set $I = \pi$, $C_0=2$ with the initial condition $u_0(x)=\sin(x)+2$, and the analytical solution of \eqref{eq:diffusion} is $u(x,t)=\mathrm{e}^{-t}\sin(x)+2.$
		
\end{example}

We first give the OVP description of \eqref{eq:diffusion} following \cite{Doi} as a preparation. The state variable $u(x,t)$ of the system (density of heat/particles at point $x$ and time $t$) is associated with the conservation equation (heat/particles conservation):
\begin{equation}\label{eq:conservation linear}
	\partial_t u + \partial_x j=0, 
	\quad j|_{\partial \Omega} = 0,
\end{equation}
with $j(x,t)$ the flux.
The free energy functional of \eqref{eq:diffusion} is given by 
$
\mathcal{E}(u) = \int_{\Omega} u \log u dx.$
By \eqref{eq:conservation linear}  and integration-by-parts, we have 
\[\dot{\mathcal{E}}(u,\partial_tu) = \int_{\Omega} (1+\log u)\partial_t u  dx = \int_{\Omega} (1+\log u)(-\partial_x j)dx = \int_{\Omega} \frac{\partial_xu}{u}jdx.
\]
Then, with the dissipation functional
\begin{equation}\label{eq:dissipation_linear}
    \Phi(u,j) = \frac{1}{2}\int_{\Omega} \frac{j^{2}}{u}dx,
\end{equation}
the Rayleighian functional \eqref{eq:reduced_form} for  (\ref{eq:diffusion}) reads
\begin{equation}\label{eq:Rayleighian linear}
	\widetilde{\mathcal{R}}(u,j) = \dot{\mathcal{E}} + \Phi
	= \int_{\Omega} \frac{\partial_{x}u}{u}jdx
	+ \frac12 \int_{\Omega} \frac{j^{2}}{u}dx .
\end{equation}
The OVP analytically gives the constitutive equation between $j$ and $u$ as
$
\delta\widetilde{\mathcal{R}}/{\delta j}=0 
\ \Rightarrow\ 
j =-\partial_{x}u.
$
Substituting this constitutive relation into the conservation equation (\ref{eq:conservation linear}) and eliminating $j$ recovers the dynamical equation
$
\partial_tu \;=\; \partial_{xx}u$.

\begin{figure}[t!]
 		\centerline{
 			\psfig{figure=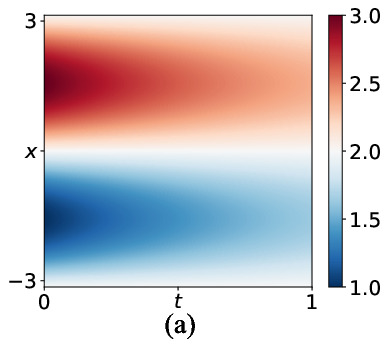,width=1.7in,height=1.5in,angle=0}
 			\hspace{-0.1in}
 			\psfig{figure=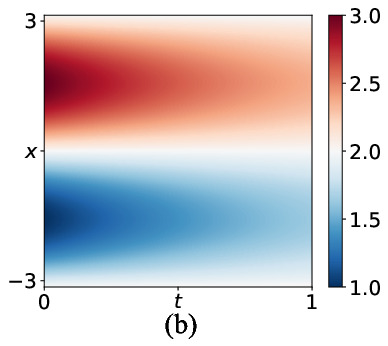,width=1.7in,height=1.5in,angle=0}
 			\hspace{-0.1in}
 			\psfig{figure=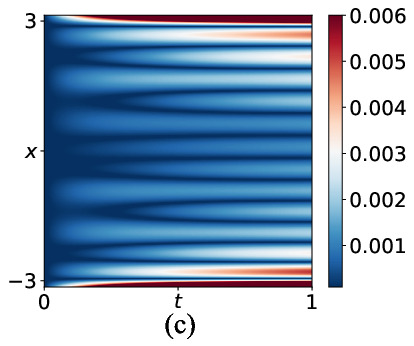,width=1.7in,height=1.5in,angle=0}
 		}
 		\centerline{
 			\psfig{figure=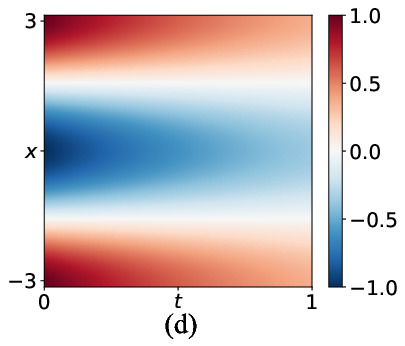,width=1.7in,height=1.5in,angle=0}
 			\hspace{-0.1in}
 			\psfig{figure=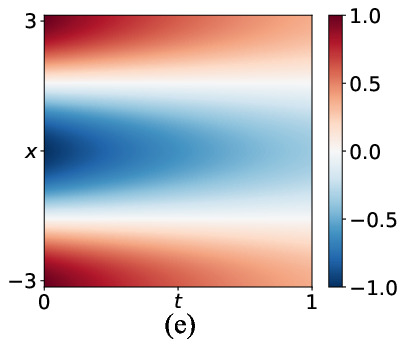,width=1.7in,height=1.5in,angle=0}
 			\hspace{-0.1in}
 			\psfig{figure=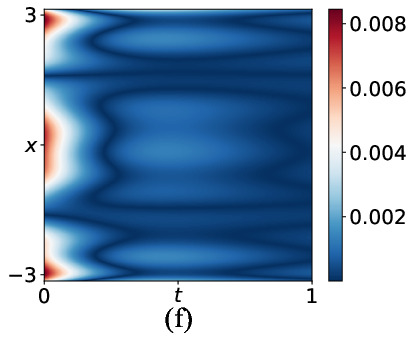,width=1.7in,height=1.5in,angle=0}
 		}
 		\caption{Results for Example \ref{ex:Linear-H} with $L=4$, $W=70$: (a) exact solution $u$; (b) numerical solution $u^n$; (c) pointwise error in $u$; (d) exact solution $j$; (e) numerical solution $j^n$; (f) pointwise error in $j$.}\label{fig:Onsager_Linear_K1}
 	\end{figure}

Now we implement the proposed Algorithm \ref{alg:workflow} to solve Example \ref{ex:Linear-H}. 
The Fourier basis is used with the truncation order $K=1$, and we generate $N_b=50$ training samples.
The input of the trunk network is the coordinates defined on a uniform spatial grid $x_k = -I+  2kI/N_x, k = 1, \dots, N_x$, where $N_x=128$  is used during training and a finer grid with $N_x=1024$ is adopted for testing error.
Set the activate function as $\sigma=\tanh$ and the hyper-parameters in (\ref{DeepONet}) as $L\in \left\{1,2,3,4 \right\}$, $W\in \left\{30,50,70 \right\}$,   $p=120$. 
By (\ref{loss_r_sum}) and discretizing the integrals in (\ref{eq:Rayleighian linear}), the practical loss function to train $\mathcal{G}_\theta$ reads
\[
\text{Loss}{(\theta)} = \frac{2I}{N_b\times N_x} \sum_{b=1}^{N_b}\sum_{k=1}^{N_x} \left[ \frac{\partial_xu^{(b)}(x_k)}{u^{(b)}(x_k)}j^{(b)}(x_k;\theta) + \frac{(j^{(b)}(x_k;\theta))^2}{2u^{(b)}(x_k)}  \right].
\]
Based on the trained operator network $\mathcal{G}_\theta$, the solutions $u$ and $j$ of (\ref{eq:diffusion}) are computed by \eqref{eq:time_d} over the entire spatiotemporal domain till $t= 1$ with $\Delta t = 0.001$. 

The exact solution and the numerical solution obtained by the DOOL method are shown in Fig. \ref{fig:Onsager_Linear_K1}, along with the corresponding pointwise error.  As observed, DOOL can accurately predict $u$ and $j$ across the spatiotemporal domain. The precision in $u$ and $j$ are similar, so we will only show the results of $u$ afterwards for brevity. Meanwhile, 
the errors of the numerical solution from DOOL under different depth and width are presented in Table \ref{tab:linear_error}. We can see that with the increase in network width and depth, the error consistently decreases, showing a better approximation of the numerical solution. This indicates the effectiveness of the network.

\begin{table*}[h!] 	
	\centering 	
	\caption{The relative spatiotemporal $L^2$-error of $u$ by DOOL with different network architecture.} 	
	\label{tab:linear_error} 	
	\begin{tabular}{c|>{\centering\arraybackslash}m{1.5cm}>{\centering\arraybackslash}m{1.5cm}>{\centering\arraybackslash}m{1.5cm}}
		\hline
		   & $W=30$        & $W=50$        & $W=70$          \\ \hline
		$L=1$   & 1.60$\,e-2$ & 9.81$\,e-3$ & 8.06$\,e-3$\\
		$L=2$    & 1.11$\,e-2$ & 1.01$\,e-2$ & 8.80$\,e-3$\\
		$L=3$    & 9.29$\,e-3$ & 6.41$\,e-3$ & 5.81$\,e-3$\\
		$L=4$    & 3.60$\,e-3$ & 4.22$\,e-3$ & 1.29$\,e-3$\\ \hline
	\end{tabular}
\end{table*}

     \begin{figure}[h!]
 		\centerline{
 			\psfig{figure=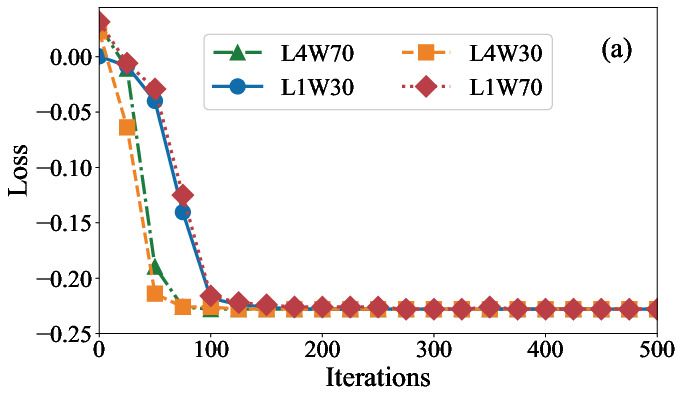,width=2.4in,height=1.5in,angle=0}
 			\hspace{0in}
 			\psfig{figure=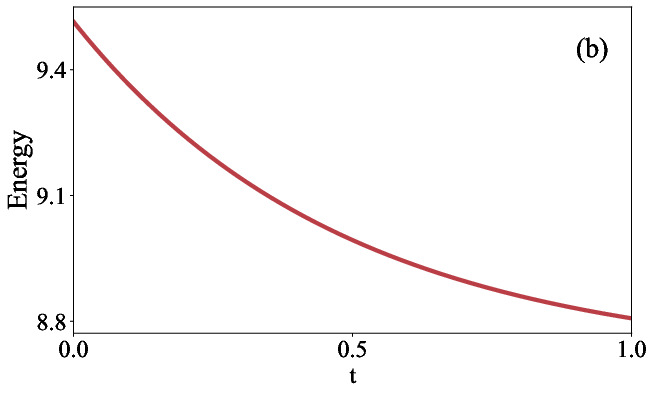,width=2.4in,height=1.5in,angle=0}
 		}
 		\caption{Results for Example \ref{ex:Linear-H} : (a) the change of Loss during the iterations; (b) energy decay over time.}\label{fig:Onsager_Linear_K1_fig2}
 \end{figure}

  Moreover, in Fig. \ref{fig:Onsager_Linear_K1_fig2}, we show the training process and the behavior of the free energy during the time stepping in DOOL. 
 As shown in Fig. \ref{fig:Onsager_Linear_K1_fig2}(a), the loss function decreases rapidly and converges with the number of training iterations increasing, indicating the good optimization characteristics of the proposed approach. Fig. \ref{fig:Onsager_Linear_K1_fig2}(b) depicts the overall energy dissipation process during the numerical simulation.


\begin{example}[With source]\label{ex:Linear-NonH}
Next, we consider a source term added to the heat equation:
\begin{equation}\label{eq:diffusion_nonhomogeneous}
	\begin{cases}
		\partial_tu(x,t) \;=\; \partial_{xx}u(x,t) + \sin(x), & x\in(-I, I),\; t>0,\\
		u(I,t)=u(-I,t)=C_0,     & t\geq0,\\
		u(x,0)=u_0(x),       & x\in[-I, I].
	\end{cases}
\end{equation}
We set $I=\pi$, $C_0=2$ and $u_0(x)=2$,  then the analytical solution of \eqref{eq:diffusion_nonhomogeneous} is $u(x,t)=(1-\mathrm{e}^{t})\sin(x)+2$.
\end{example}

\begin{figure}[h!]
	\centerline{
		\psfig{figure=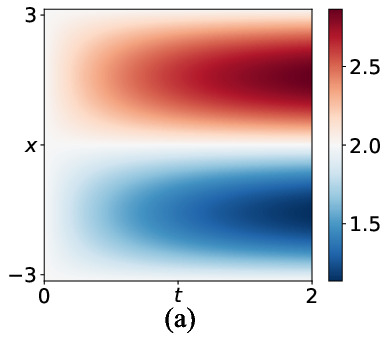,width=1.7in,height=1.5in,angle=0}
		\hspace{-0.1in}
		\psfig{figure=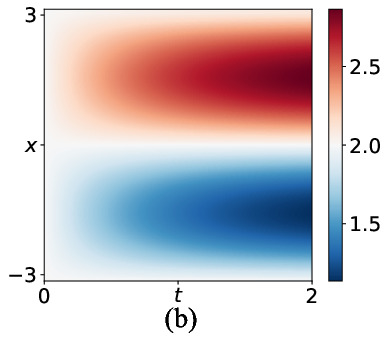,width=1.7in,height=1.5in,angle=0}
		\hspace{-0.1in}
		\psfig{figure=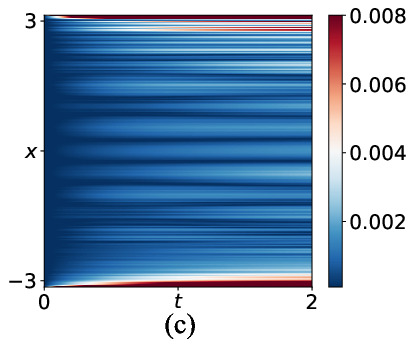,width=1.7in,height=1.5in,angle=0}
	}
	\caption{Results for Example \ref{ex:Linear-NonH}: (a) exact solution $u$; (b) numerical solution $u^n$; (c) pointwise error in $u$.}\label{fig:Onsager_Linear_Nonhomogeneous}
\end{figure}

In this case, the free  energy functional is identical to that of  \eqref{eq:diffusion}, and we define the dissipation functional as
\[
\Phi(u,j) = \int_{\Omega} \left(\frac{j^{2}}{2u} - \frac{\cos(x)}{u}j\right) dx.
\]
Repeating the derivation process of Example \ref{ex:Linear-H}, the Rayleighian is given by
\begin{equation*}
	\widetilde{\mathcal{R}}(u,j) = \dot{\mathcal{E}} + \Phi
	= \int_{\Omega} \frac{\partial_{x}u}{u}jdx +
	\int_{\Omega} \left(\frac{j^{2}}{2u} - \frac{\cos(x)}{u}j\right) dx.
\end{equation*}
According to the OVP, ${\delta\widetilde{\mathcal{R}}}/{\delta j} = 0$ resulting in $j = -{\partial_{x}u}  + \cos(x).$
Substituting this constitutive equation into (\ref{eq:conservation linear}) yields \eqref{eq:diffusion_nonhomogeneous}.

For the numerical test, the training is performed using $N_b = 20$ samples with hyper-parameters $L=4$, $W=70$, $p=180$, and all other settings follow Example \ref{ex:Linear-H}.
The exact solution and the numerical solutions obtained by our method are shown in Fig. \ref{fig:Onsager_Linear_Nonhomogeneous} up to $t=2$, along with the corresponding pointwise error. The relative spatiotemporal $L^2$-error of $u$ in this case is $3.52\times10^{-3}$. This results validate DOOL on this problem.

\begin{example}[Fokker--Planck equation]\label{ex:fp}
Consider the Fokker--Planck equation:
\begin{equation}\label{eq:fp}
	\begin{cases}
		\partial_t u(x,t) = \beta^{-1} \partial_{xx} u(x,t) + \partial_x\left(  u(x,t) \cdot\partial_x V(x)  \right), & x \in (-I, I),\; t>0, \\
		u(-I,t)=u(I,t),\  \partial_{x}u(-I,t)=\partial_{x}u(I,t), & t\geq0, \\
		u(x,0)=u_0(x), & x\in [-I, I],
	\end{cases}
\end{equation}
where $u$ denotes the particle density evolving under diffusion and the drag force from a potential field $V(x)$. Let $\beta = 2$, $I=5$, $V(x) = \frac{1}{2}x^2$ and $u_0(x) = \frac{1}{\sqrt{2\pi}}\mathrm{e}^{-{x^2}/{2}}$, then the analytical solution of \eqref{eq:fp} is $u(x,t) = \frac{1}{\sqrt{2\pi\sigma_t^2}}\mathrm{e}^{-{x^2}/{2\sigma_t^2}}$ with $\sigma_t^2 = \beta^{-1}(1-\mathrm{e}^{-2t})+\mathrm{e}^{-2t}$.
\end{example}

In this system, $u$ is conserved and satisfies \eqref{eq:conservation linear}. The free energy functional following \cite{Onsagerchen} is defined by
\[
\mathcal{E}(u) = \int_{\Omega} \beta^{-1} \left(u \log u + u V\right) \, dx.
\]
Thus, 
\begin{align*}
\dot{\mathcal{E}}(u,\partial_tu) 
&= \int_{\Omega} \left( \beta^{-1} (1 + \log u)  +  V\right)\partial_t u  dx \\
&= \int_{\Omega} \left( \beta^{-1} (1 + \log u)  +  V\right)(-\partial_x j)dx 
= \int_{\Omega} \left( \beta^{-1} \frac{\partial_x u}{u} + \partial_x V \right)  j\,dx.
\end{align*}
With the dissipation term given as \eqref{eq:dissipation_linear}, 
the Rayleighian functional reads
\begin{align}\label{eq:R_FP}
	\widetilde{\mathcal{R}}(u,j)
	&= 
	 \int_{\Omega} \left( \beta^{-1} \frac{\partial_x u}{u} + \partial_x V \right)  j \, dx
	+ \int_{\Omega} \frac{|j|^2}{2u}\, dx.
\end{align}
The constitutive relationship from the OVP
$   j=-\beta^{-1} {\partial_x u} - u\partial_x V$ 
together with \eqref{eq:conservation linear} leads to the Fokker-Planck equation \eqref{eq:fp}.

\begin{figure}[h!]
		\centerline{
		\psfig{figure=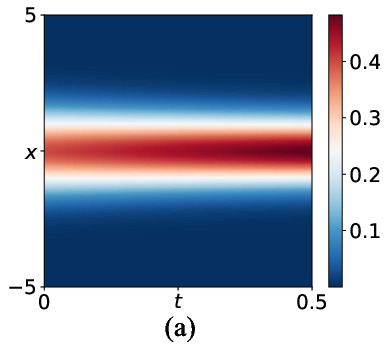,width=1.7in,height=1.5in,angle=0}
		\hspace{-0.1in}
		\psfig{figure=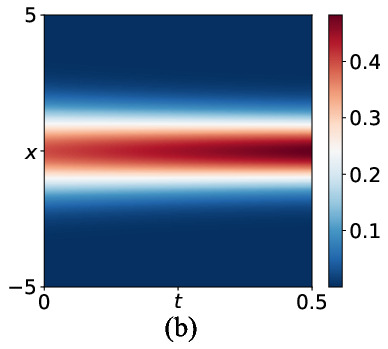,width=1.7in,height=1.5in,angle=0}
		\hspace{-0.1in}
		\psfig{figure=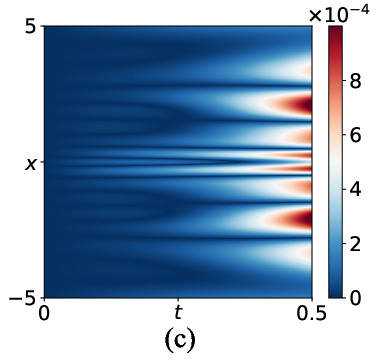,width=1.7in,height=1.5in,angle=0}
	}
	\caption{Results for the Fokker-Planck equation in Example \ref{ex:fp}: (a) exact solution $u$; (b) numerical solution $u^n$; (c) pointwise error in $u$.}\label{fig:Onsager_FP}
\end{figure}

In this problem, we consider the Hermite basis, the branch network takes the coefficient vector of the state field on the Hermite basis as input with the truncation order $K=5$, and we generate training samples $N_b=500$.
The input of the trunk network is the uniform spatial grid with $N_x=400$ used during training, while a finer grid with $N_x=2000$ is used for the error testing.
The hyper-parameters are set as: $\Delta t=0.001$, $L = 4$, $W = 70$ and $p=120$, with the activate function $\sigma=\tanh$ employed.
We show the solutions and errors till $t=0.5$ with $\Delta t = 0.001$ in Fig. \ref{fig:Onsager_FP}, and the relative spatiotemporal $L^2$-error of $u$ is $1.16\times10^{-3}$. The results again prove the validity of DOOL. 

\begin{remark}
    The possible small value or zero of $u$ in the denominator of the Rayleighian functional, e.g., \eqref{eq:R_FP}, may cause numerical instability.
    This can be overcome simply by considering a shifted variable $u\to u + C$ (for some constant $C>0$), leading to an equivalent problem of the same form because of linearity. 
\end{remark}

\begin{example}[Cahn–Hilliard equation]\label{ex:CH}
Consider the Cahn–Hilliard equation:
\begin{equation}\label{eq:cahn-hilliard}
	\begin{cases}
		\partial_{t}u(\bx,t) \;=\; \Delta
		\!\bigl(-\gamma_1 \Delta u(\bx,t) + \gamma_2\left(u(\bx,t)^3-u(\bx,t)\right)\bigr),
		& \bx\in\Omega,\; t>0,\\[4pt]
		u(-\bx,t)=u(\bx,t),\  \nabla u(-\bx,t)=\nabla u(\bx,t)
		&  \bx \in \partial\Omega,\; t\geq0,\\[4pt]
		u(\bx,0)=u_0(\bx),
		& \bx\in \overline{\Omega},
	\end{cases}
\end{equation}
which is a nonlinear model for chemical diffusion and reaction. Here $\gamma_1,\gamma_2\in\mathbb{R}$ are given and $\Omega\subset \mathbb{R}^d$.
\end{example}

\begin{figure}[h!]
	\centerline{
		\psfig{figure=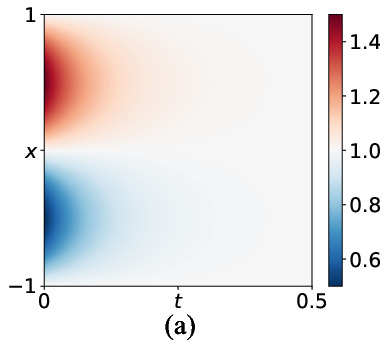,width=1.7in,height=1.5in,angle=0}
		\hspace{-0.1in}
		\psfig{figure=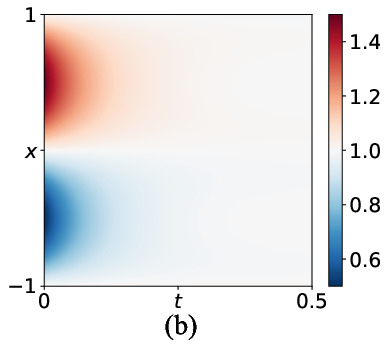,width=1.7in,height=1.5in,angle=0}
		\hspace{-0.1in}
		\psfig{figure=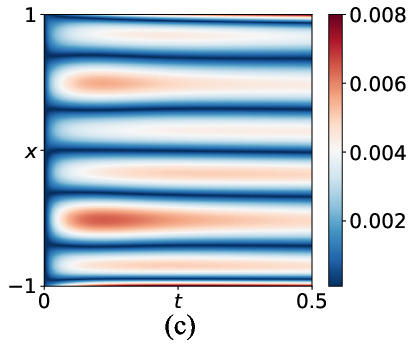,width=1.7in,height=1.5in,angle=0}
	}

    \centerline{
		\psfig{figure=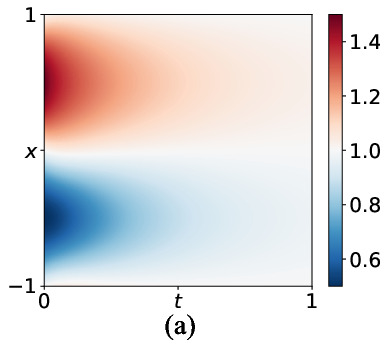,width=1.7in,height=1.5in,angle=0}
		\hspace{-0.1in}
		\psfig{figure=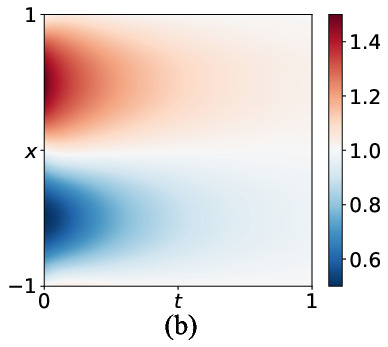,width=1.7in,height=1.5in,angle=0}
		\hspace{-0.1in}
		\psfig{figure=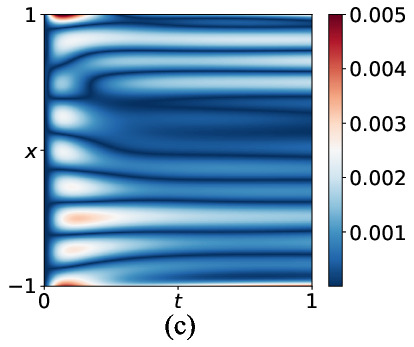,width=1.7in,height=1.5in,angle=0}
	}
	\caption{Results for the 1D Cahn–Hilliard equation with $\gamma_1 = 0.1$ (1st row) and $\gamma_1 = 0.01$ (2nd row) in Example \ref{ex:CH}: (a) exact solution $u$; (b) numerical solution $u^{n}$; (c) pointwise error of $u$.}\label{fig:Onsager_CH_1}
\end{figure}

Again, we go through the OVP as a benchmark. In this system, the unknown state  $u$ is conserved and satisfies  
\begin{equation}\label{eq:conservation ch 2D}
	\partial_t u + \nabla \cdot \boldsymbol{j} = 0,
	\qquad
	{\boldsymbol{j} \cdot \boldsymbol{n}|_{\partial\Omega}=0,}
\end{equation}
where $\boldsymbol{j}=(j_1,\dots, j_d)^\top \,\text{:}\, \Omega \rightarrow\mathbb{R}^d$ with each  $j_k$, $k=1,\dots,d$ representing the flux along the $k$-th spatial direction.
The free energy functional defined by \cite{Onsagerchen} reads
\[
\mathcal{E}(u)
=\int_{\Omega}
\Bigl(\frac{\gamma_1}{2}(\nabla u)^{2}
+\frac{\gamma_2}{4}\bigl(u^{2}-1\bigr)^{2}\Bigr)\,d\bx,
\]
and direct calculations yield
\begin{align*}
\dot{\mathcal{E}}(u,\partial_t u) 
&= \int_{\Omega} \left(-\gamma_1 \Delta u + \gamma_2(u^{3}-u)\right)\partial_t u  d\bx \\
&= \int_{\Omega} \left(-\gamma_1 \Delta u + \gamma_2(u^{3}-u)\right)(-\nabla \cdot \boldsymbol{j}) d\bx \\
&= \int_{\Omega} \nabla\left(-\gamma_1 \Delta u + \gamma_2(u^{3}-u)\right)\cdot\boldsymbol{j}\,d\bx.
\end{align*}
In this scenario, the dissipation functional is defined as
$
\Phi(\boldsymbol{j})=\frac{1}{2}\int_{\Omega} \boldsymbol{j}^\top\boldsymbol{j}\,d\bx.
$
Thus, we can find the Rayleighian functional as
\begin{align}\label{eq:Rayleighian ch}
	\widetilde{\mathcal{R}}(u,\boldsymbol{j})  
	&=\int_{\Omega} \nabla\left(-\gamma_1 \Delta u + \gamma_2(u^{3}-u)\right) \cdot \boldsymbol{j}\,d\bx
	+\frac{1}{2}\int_{\Omega} \boldsymbol{j}^\top\boldsymbol{j}\,d\bx.
\end{align}
Setting $\delta\widetilde{\mathcal{R}}/\delta \boldsymbol{j} = 0$ yields 
 $   \boldsymbol{j} = \nabla\left(\gamma_1 \Delta u - \gamma_2(u^{3}-u)\right).$
Substituting it into \eqref{eq:conservation ch 2D} gives
 the Cahn–Hilliard equation \eqref{eq:cahn-hilliard}.

   \begin{figure}[h!]
 		\centerline{
 			\psfig{figure=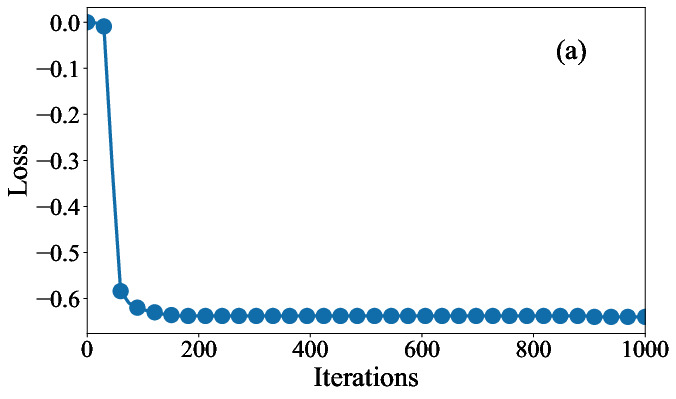,width=2.4in,height=1.5in,angle=0}
 			\hspace{0in}
 			\psfig{figure=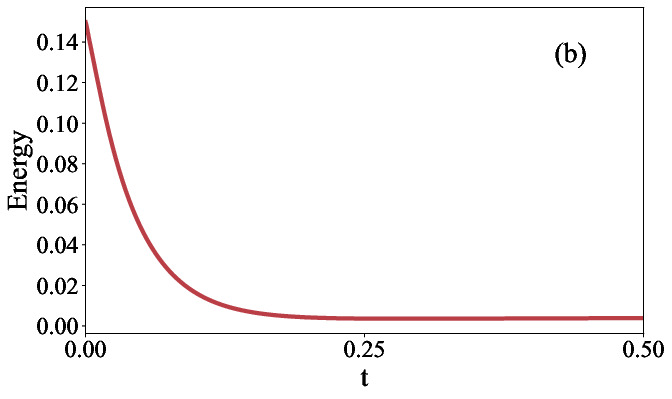,width=2.4in,height=1.5in,angle=0}
 		}
 		\caption{Results for Example \ref{ex:CH} with $\gamma_1=\gamma_2 = 0.1$: (a) the change of Loss during the iterations; (b) energy decay over time.}\label{fig:CH_loss_energy}
 \end{figure}

\begin{figure}[h!]
	\centerline{
		\psfig{figure=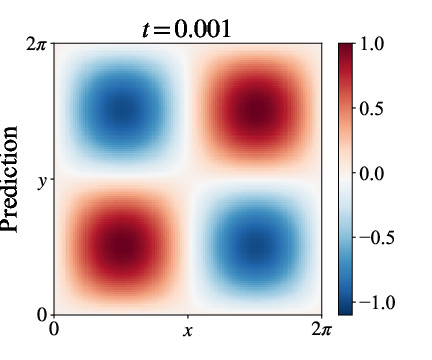,width=1.5in,height=1.3in,angle=0}
		\hspace{-0.2in}
		\psfig{figure=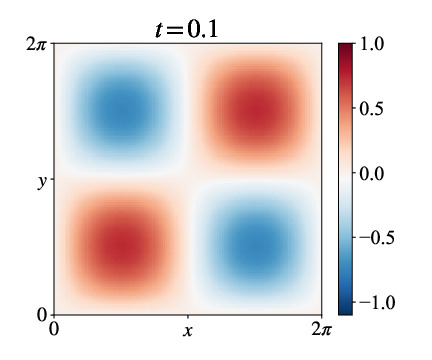,width=1.5in,height=1.3in,angle=0}
		\hspace{-0.2in}
		\psfig{figure=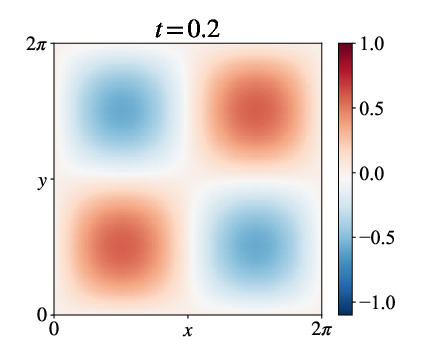,width=1.5in,height=1.3in,angle=0}
	}

    \vspace{-0.1in}

	\centerline{
		\psfig{figure=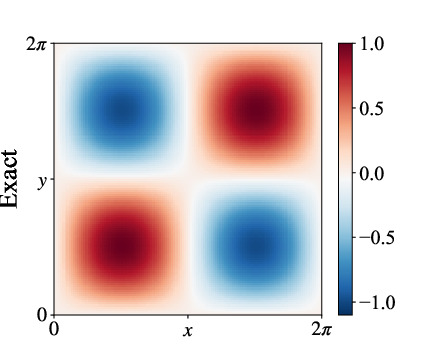,width=1.5in,height=1.3in,angle=0}
		\hspace{-0.2in}
		\psfig{figure=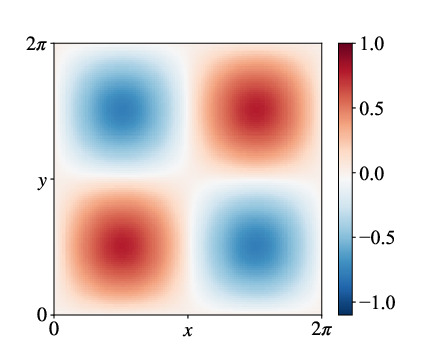,width=1.5in,height=1.3in,angle=0}
		\hspace{-0.2in}
		\psfig{figure=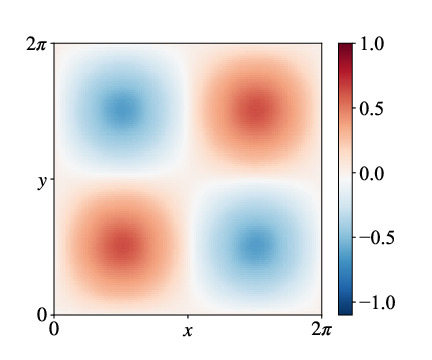,width=1.5in,height=1.3in,angle=0}
	}

    \vspace{-0.1in}
    
	\centerline{
		\psfig{figure=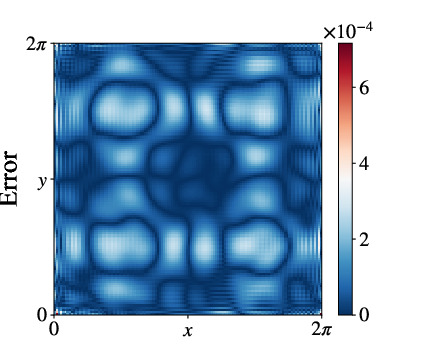,width=1.5in,height=1.3in,angle=0}
		\hspace{-0.2in}
		\psfig{figure=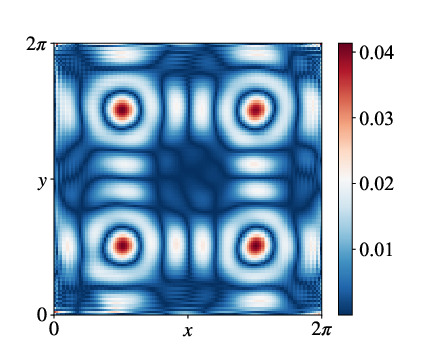,width=1.5in,height=1.3in,angle=0}
		\hspace{-0.2in}
		\psfig{figure=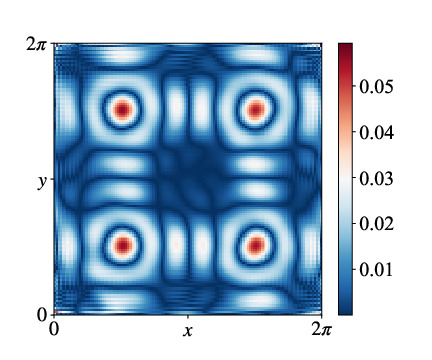,width=1.5in,height=1.3in,angle=0}
	}
	\caption{Results for the 2D Cahn–Hilliard equation: numerical solutions $u$ (1st row); exact solutions $u^{n}$ (2nd row); pointwise error of $u$ (3rd row).}\label{fig:Onsager_CH_2D}
\end{figure}

We begin with the 1D case ($d=1$) for testing. Set $\gamma_1 =0.1$, $\gamma_2 =0.1$, $\Omega = (-1, 1)$, and the initial condition 
$u_0(x)=\tfrac{1}{2}\sin(\pi x)+1$. The reference solution is obtained by a classical solver \cite{ETD}.
Applying DOOL, 
we take the Fourier basis with $K=2$.
Set the number of samples in the training to $N_b = 500$ and
take the activate function as $\sigma=\sin$. The hyper-parameters are taken as $L = 4$, $W = 90$, $p=180$, $\Delta t=0.001$.
The exact and numerical solutions till $t=0.5$ (where steady state is reached) together with the pointwise error are shown in Fig. \ref{fig:Onsager_CH_1}. The relative $L^2$-error in $u$ is $3.30\times10^{-3}$. 
{Fig. \ref{fig:CH_loss_energy} presents the training process and the evolution of the free energy during the time stepping.}
As can be seen, the results illustrate that DOOL can also perform well on the nonlinear problem. The observed numerical energy dissipation, i.e., Fig. \ref{fig:CH_loss_energy}(b) and Fig. \ref{fig:Onsager_Linear_K1_fig2}(b), is not some coincidence, based on numerical results not shown for brevity here and after. We will address this behavior in the appendix.

Furthermore, we consider a smaller value for the parameter $\gamma_1$ by taking $\gamma_1=0.01$ with the others unchanged. This creates a sharper interface (larger gradient) between the two phases, and we aim to demonstrate that DOOL is feasible for such a situation. 
With a minor adjustment: $K=3,\, N_b=250,\,W=70$, 
the results till $t=1$ are included in Fig. \ref{fig:Onsager_CH_1}, where the relative $L^2$-error is $1.07\times10^{-3}$ for $u$. The results are similar, leading to the same conclusion.

Last but not least, we present a two-dimensional (2D) test of the example, with the reference solution obtained similarly as before. Take $\gamma_1 =\gamma_2 =1$, $\Omega = (0, 2\pi) \times (0, 2\pi)$, 
and $u_0(x,y) = \sin(x)\sin(y)$. 
Set $K=1$ under the Fourier basis and use $N_b=100$ samples for training.
A uniform grid with $N_x=N_y=128$ in the $x$ and $y$ directions is used as the input for the trunk network during training and $N_x=N_y=400$ is adopted for testing error.
Set $\sigma=\sin$ and
$K=1$, $L = 4$, $W = 70$, $p=180$, $\Delta t=0.001$.
We show the solutions and errors at
$t=0.001$, $t=0.1$ and $t=0.2$ in Fig. \ref{fig:Onsager_CH_2D}. 
The results indicate the potential of DOOL to work for high-dimensional problems.

\section{Generalization  and comparison}\label{sec:compare}
In this section, we will first illustrate the generalization ability of DOOL in terms of the initial condition of a PDE model and conduct systematical comparisons with the existing supervising-mannered DeepONet \cite{DeepONet}, justifying the efficiency of the proposed approach. Then, we will investigate the generalization capability of DOOL with respect to more model parameters, and compare it with the existing MIONet approach \cite{Jin} for multiple-input operator learning.

\subsection{Comparison with DeepONet}
We take the 1D Cahn-Hilliard equation from Example \ref{ex:CH} as the test case for the generalization capability of DOOL for the initial condition in (\ref{eq:cahn-hilliard}).
For comparisons, in the following, let us refer to DeepONet as the standard approach from \cite{DeepONet} using supervised learning \eqref{eq:loss_deeponet}.

In (\ref{eq:cahn-hilliard}), the parameters are set as $\gamma_1 =0.01$, $\gamma_2 = 0.1$, $\Omega = (-1, 1)$.
All DOOL training settings are consistent with those specified in Example \ref{ex:CH}.
To ensure a fair comparison with DeepONet, both approaches employ identical network architectures and spatial discretization with $N_x=128$ in training. 
The key difference in training lies in: DeepONet requires uniformly discrete grid points in the temporal domain $[0,1]$ with $N_t = 50$, which are combined with spatial grid points to form the input of the trunk network.
The training settings of both methods are presented in Table \ref{tab:com_deeponet}, showing that DOOL requires less training time than DeepONet under equivalent training iterations.
This advantage comes from the fact that DOOL's training process does not require input of time coordinates, thereby reducing the amount of data.

\begin{table}[h!]
	\centering
	\caption{Training settings of DOOL and DeepONet.}
	\begin{tabular}{ccccccc}
		\hline
		\rule{0pt}{10pt}
		Setting & L & W & p & Parameters & Epochs & Training time(s) \\
		\hline
		\rule{0pt}{10pt}
		DeepONet & 4 & 70 & 180 & 56080 & 200000  & 832.82 \\
		DOOL & 4 & 70 & 180 & 56010 & 200000  & 776.81 \\
		\hline
		
	\end{tabular}
	\label{tab:com_deeponet}
\end{table}

\begin{figure}[h!]
	\centerline{
		\psfig{figure=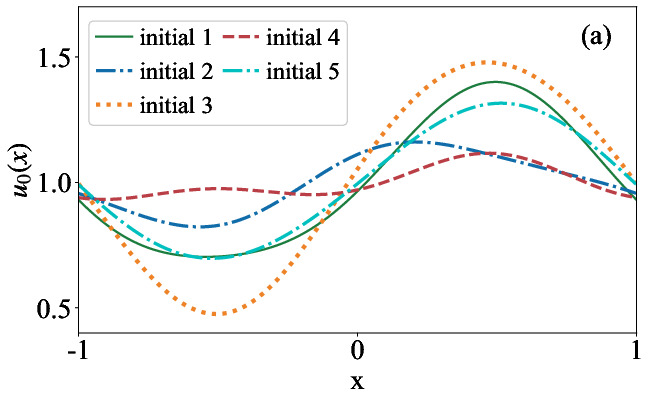,width=2.4in,height=1.5in,angle=0}
		\hspace{0in}
		\psfig{figure=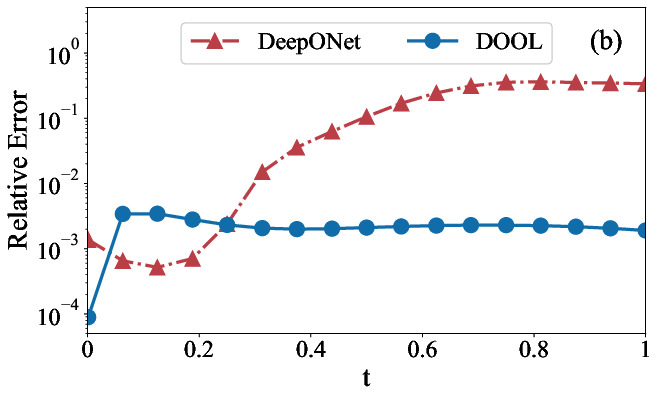,width=2.4in,height=1.5in,angle=0}
	}

	\caption{(a): test set of initial conditions; (b): time extrapolation comparison under the initial condition $1$.}\label{fig:diff_u0}
\end{figure}

To evaluate the generalization performance of initial value, we randomly select 5 untrained initial values to form the test set.
The test set of initial values are visualized in Fig. \ref{fig:diff_u0}(a).
For these initial conditions, we numerically solve the PDE using both the DOOL and DeepONet on a finer spatiotemporal grid ($N_x=1000$, $N_t=900$) on $[-1,1]\times[0,1]$. Their relative spatiotemporal $L^2$-errors are compared in Table \ref{tab:comp_ini_error_deeponet}.
As demonstrated in Table \ref{tab:comp_ini_error_deeponet}, our proposed unsupervised training method achieves a precision comparable to the standard supervised DeepONet approach.

\begin{table}[h!]
	\centering
	\caption{Accuracy comparison of DOOL and DeepONet for initial value generalization.}
	\begin{tabular}{cccccc}
		\hline
		\rule{0pt}{10pt}
		Initial & 1 & 2 & 3 & 4 & 5 \\
		\hline
		\rule{0pt}{10pt}
		DeepONet & 1.89$\,e-3$ & 1.56$\,e-3$ & 1.83$\,e-3$ & 1.57$\,e-3$ & 2.06$\,e-3$ \\
		DOOL & 2.38$\,e-3$ & 2.31$\,e-3$ & 2.20$\,e-3$ & 1.55$\,e-3$ & 1.27$\,e-3$ \\
		\hline
		
	\end{tabular}
	\label{tab:comp_ini_error_deeponet}
\end{table}

\begin{table}[h!]
	\centering
	\caption{Comparison of DOOL and DeepONet in temporal extrapolation.}
	\label{tab:comp_error_deeponet ex}
	\begin{tabular}{ccccccc}
		\toprule
		\multirow{2}{*}{\shortstack{\\ \\ Time \\ Interval}} & \multirow{2}{*}{\shortstack{\\ \\Method}} & \multicolumn{5}{c}{Initial} \\
		\cmidrule(lr){3-7}
		& & {1} & {2} & {3} & {4} & {5} \\
		\midrule
		
		\multirow{2}{*}{$[0,\,0.5]$}  
		& DeepONet &  3.50$\,e-2$ & 1.99$\,e-2$ & 2.17$\,e-2$ & 3.04$\,e-2$ & 2.93$\,e-2$ \\
		& DOOL     & 2.54$\,e-3$ & 2.09$\,e-3$ & 2.03$\,e-3$ & 1.73$\,e-3$ & 1.48$\,e-3$ \\
		
		\midrule
		
		\multirow{2}{*}{$[0,\,1]$}    
		& DeepONet &  2.16$\,e-1$ & 8.25$\,e-2$ & 1.05$\,e-1$ & 2.00$\,e-1$ & 1.47$\,e-1$ \\
		& DOOL     & 2.38$\,e-3$ & 2.31$\,e-3$ & 2.20$\,e-3$ & 1.55$\,e-3$ & 1.27$\,e-3$ \\
		
		\bottomrule
	\end{tabular}
\end{table}

Next, we demonstrate another major advantage of DOOL over DeepONet, which is its temporal extrapolation capability.  
For training, now DeepONet inputs uniform grid points in a prescribed time interval $[0,\,0.2]$ with $N_t = 20$, while DOOL maintains its architecture without time input. 
All other hyper-parameters remain identical as in Table \ref{tab:com_deeponet}. 
Using the initial condition test set from Fig. \ref{fig:diff_u0}(a), we numerically solve the model on an extended time interval $[0,\,0.5]$ or $[0,\,1]$. The relative errors of the two methods with the spatiotemporal discrete setting of $N_x=1000,\,\Delta t = 0.001$
are presented in Table \ref{tab:comp_error_deeponet ex}. We can see that in both short-term  ($t= 0.5$) and long-term  ($t=1$) predictions, DOOL achieves significantly higher accuracy than DeepONet. The result demonstrates DOOL's superior performance in the temporal extrapolation task. Furthermore, we plot the error from the initial condition $1$ against $t$ in Fig. \ref{fig:diff_u0}(b). It shows that as the extrapolation time increases, DOOL will be more effective than the standard DeepONet approach.

\subsection{Multiple-input case and comparison with MIONet}\label{sec:gene_muti}
We further consider the possible need for generalization with respect to multiple parameters/functions in a PDE model.  This in fact is quite  straightforward under DOOL by adding additional branches to (\ref{DeepONet}), as done in the Multiple-Input Operator Network (MIONet) \cite{Jin}. 

\begin{figure}[h!]
	\centerline{
		\psfig{figure=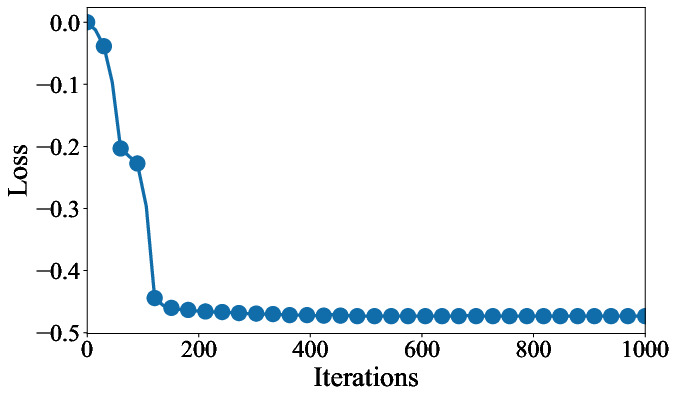,width=2.8in,height=1.5in,angle=0}
	}

	\caption{Loss of DOOL during the training in the multiple-input case.}\label{fig:MIO_loss}
\end{figure}

Considering still the 1D Cahn–Hilliard equation \eqref{eq:cahn-hilliard} with $\Omega = (-1, 1)$ and $\gamma_2=0.1$, 
the current implementation simultaneously generalizes the initial condition and the coefficient $\gamma_1$. 
In this case, the network $\mathcal{G}_\theta:(u^n,\gamma_1) \mapsto j^n$ is used to approximate the constitutive relation
$\mathcal{G} : (u,\gamma_1) \mapsto j$, i.e., $j
= \partial_x\bigl(\gamma_1 \partial_{xx}u - \gamma_2(u^{3}-u)\bigr)$, by minimizing the Rayleighian function \eqref{eq:Rayleighian ch}. 
Consequently, two branch networks are used to receive ${u}$ and the parameter $\gamma_1$ as inputs, respectively.
For the first branch network inputting $u$, the Fourier basis with truncation order $K=3$ and $N_b = 120$ training samples.
For the second branch network inputting $\gamma_1\in [0.01,0.1]$, we independently sample $N_l = 10$ training samples from a uniform distribution over the interval.
The input of the trunk network is the uniform spatial grid with $N_x=128$ used during training, and $N_x=1024, \Delta t =0.001$ is adopted for testing error.
The hyper-parameters are set as: $L = 4$, $W = 70$ and $p=120$, with the activate function $\sigma=\sin$ employed.
By (\ref{loss_r_sum}) and discretizing the integrals in \eqref{eq:Rayleighian ch}, the practical loss function for DOOL reads
\[
\text{Loss}{(\theta)} = \frac{2I}{N_b\times N_l\times N_x} \sum_{b=1}^{N_b}\sum_{l=1}^{N_l}\sum_{k=1}^{N_x} \left[ F(\gamma_1^{(l)} ,u^{(b)}(x_k))j^{(b)}(x_k;\theta) + \frac{(j^{(b)}(x_k;\theta))^2}{2}  
\right],
\]
where $F(\gamma_1,u^n) = \partial_{x}\left(-\gamma_1 \partial_{xx}u^n + \gamma_2((u^n)^{3}-u^n)\right)$. 
{Fig. \ref{fig:MIO_loss} shows that the loss function decreases rapidly and converges with the number of training iterations increasing, indicating that DOOL still maintains the good optimization characteristics in the multiple-input case.} 
This in fact illustrates the natural extensibility of DOOL for systems with multiple inputs, where operator training can be done simply by introducing multiple-layer summation terms into the loss function in the uniform manner, all while maintaining an unsupervised training paradigm.

After training, we test some $\gamma_1$ outside the training set with the same initial condition (1D case) from Example \ref{ex:CH}. 
Table \ref{tab:mionet_ch} shows the corresponding relative spatiotemporal $L^2$-error of $u$ and $j$ with $t\in[0,1]$. 
We can see that DOOL successfully achieves generalization across different $\gamma_1$ values while maintaining a good prediction accuracy. This
demonstrates that the DOOL method works well for multiple inputs.

	\begin{table}[t!]
	\centering
	\caption{Test errors of DOOL under different $\gamma_1$ values.}
	\begin{tabular}{cccccc}
		\hline
		\rule{0pt}{10pt}
		$\gamma_1$ & 0.01 & 0.02 & 0.03 & 0.04 & 0.05 \\
		\hline
		\rule{0pt}{10pt}
		Error in $u$ & 8.35$\,e-3$ & 7.69$\,e-3$ & 1.35$\,e-2$ & 2.37$\,e-2$ & 3.38$\,e-2$ \\
		Error in $j$ & 8.51$\,e-3$ & 6.07$\,e-3$ & 6.44$\,e-3$ & 6.50$\,e-3$ & 6.41$\,e-3$ \\
		
		\hline
		\rule{0pt}{10pt}
		$\gamma_1$ & 0.06 & 0.07 & 0.08 & 0.09 & 0.1  \\
		\hline
		\rule{0pt}{10pt}
		Error in $u$ & 4.02$\,e-2$ & 3.99$\,e-2$ & 3.07$\,e-2$ & 1.43$\,e-2$ & 3.06$\,e-2$ \\
		Error in $j$ & 6.48$\,e-3$ & 6.79$\,e-3$ & 7.28$\,e-3$ & 7.97$\,e-3$ & 9.06$\,e-3$ \\
		\hline
		
	\end{tabular}
    \label{tab:mionet_ch}
\end{table}

\begin{table}[t!]
\centering
\caption{Comparison of MIONet and  DOOL for initial value and $\gamma_1$ generalization.}
\begin{tabular}{ccccccc}
\toprule
Initial & Method & $\gamma_1=0.02$ & $\gamma_1=0.04$ & $\gamma_1=0.06$ & $\gamma_1=0.08$ & $\gamma_1=0.1$ \\
\midrule
\multirow{2}{*}{1}  & MIONet & $2.22\,e-1$ & $2.49\,e-1$ & $2.64\,e-1$ & $2.75\,e-1$ & $2.84\,e-1$ \\
                    & DOOL   & $6.96\,e-4$ & $3.18\,e-3$ & $3.84\,e-3$ & $3.52\,e-3$ & $6.78\,e-3$ \\
\cline{2-7}
\multirow{2}{*}{2}  & MIONet & $7.37\,e-2$ & $5.22\,e-2$ & $4.21\,e-2$ & $3.72\,e-2$ & $3.50\,e-2$ \\
                    & DOOL   & $2.57\,e-3$ & $2.60\,e-3$ & $3.47\,e-3$ & $5.69\,e-3$ & $1.22\,e-2$ \\
\cline{2-7}
\multirow{2}{*}{3}  & MIONet & $4.94\,e-2$ & $4.29\,e-2$ & $4.89\,e-2$ & $5.59\,e-2$ & $6.20\,e-2$ \\
                    & DOOL   & $4.08\,e-3$ & $4.85\,e-3$ & $5.09\,e-3$ & $4.05\,e-3$ & $5.36\,e-3$ \\
\cline{2-7}
\multirow{2}{*}{4}  & MIONet & $8.43\,e-2$ & $7.81\,e-2$ & $7.93\,e-2$ & $8.22\,e-2$ & $8.51\,e-2$ \\
                    & DOOL   & $1.41\,e-3$ & $2.81\,e-3$ & $3.91\,e-3$ & $5.42\,e-3$ & $1.11\,e-2$ \\
\cline{2-7}
\multirow{2}{*}{5}  & MIONet & $4.15\,e-2$ & $5.10\,e-2$ & $6.44\,e-2$ & $7.49\,e-2$ & $8.30\,e-2$ \\
                    & DOOL   & $9.62\,e-4$ & $3.49\,e-3$ & $4.32\,e-3$ & $3.87\,e-3$ & $6.41\,e-3$ \\
\bottomrule
\end{tabular}
\label{tab:comp_mionet}
\end{table}

Then, we compare the performance of the proposed DOOL approach with the original supervised learning manner \cite{Jin} (referred to later as MIONet). It is already obvious that the zero need of labeled data in DOOL framework saves much more computational costs for the multiple-input case than MIONet.  
Now we give a quantitative examination for its generalization capability by the same numerical example. 
All MIONet settings are set the same as DOOL, except that MIONet requires uniformly discrete grid points in the temporal domain with $N_t = 1000$, which are combined with spatial grid points to form the input of the trunk network.
The test set includes the initial conditions from Fig. \ref{fig:diff_u0}(a) and $\gamma_1\in\left\{0.02, 0.04, 0.6, 0.08, 0.1\right\}$. 
Table \ref{tab:comp_mionet} provides a quantitative comparison of numerical accuracy for the two methods on the test set, measuring the relative spatiotemporal $L^2$-error of $u$.
The results show that DOOL indeed exhibits higher accuracy, which implies superb generalization capability.

\section{Extension of the approach}\label{sec:extension}
This section gives some applications and extensions of the proposed DOOL approach. It contains the application for the inverse problem, the extension to systems for nonconserved parameters, and the extension for dissipative second-order wave-type equations where the analogy of DOOL based on the principle of least action will be developed.

\subsection{Parameter inversion}
A well-trained operator network can be applied to solve inverse problems directly. For illustration, here we consider a model parameter identification problem for the Cahn-Hilliard equation from Section \ref{sec:gene_muti}.

Given observational data $u^{*}(\mathbf{x}, t)$ collected at spatiotemporal coordinates $\{(\mathbf{x}_k, t_n)\}_{k=1,n=1}^{N_x, N_t}$, the objective is to recover the unknown physical parameter $\gamma_1$ in \eqref{eq:cahn-hilliard} that minimizes the difference between the observed data $u^{*}$ and the model prediction $u_{\gamma_1}$ given by DOOL with initial value $u^{*}(\mathbf{x}, t_0)$.
This inverse problem is formulated as the optimization task:
\begin{equation}\label{eq:inverse problem}
    \min_{\gamma_1 \in \Gamma}\sum_{k=1}^{N_x} \sum_{n=1}^{N_t} \left| u^{*}(\mathbf{x}_k, t_n) -u_{\gamma_1}(\mathbf{x}_k, t_n)\right|^2, \quad  \Gamma = [\gamma_{\min}, \gamma_{\max}].
\end{equation}
As the trained operator is able to accurately and efficiently predict the solution with the generalization capability for $\gamma_1$, the value of $u_{\gamma_1}$ is readily available through forward propagation for each $\gamma_1$ given in \eqref{eq:cahn-hilliard}. Therefore, it is very convenient to directly utilize some search algorithm, e.g., the golden section search for solving \eqref{eq:inverse problem}. 

\begin{table}[h!]
\centering
\caption{Error of $\gamma_1$ by golden section search.}
\begin{tabular}{cccccc}
\toprule
Initial & $\gamma_1=0.02$ & $\gamma_1=0.04$ & $\gamma_1=0.06$ & $\gamma_1=0.08$ & $\gamma_1=0.1$ \\
\midrule
1  & $2.10\,e-3$ & $5.02\,e-3$ & $2.70\,e-3$ & $1.55\,e-2$ & $3.07\,e-2$ \\
2  & $1.11\,e-3$ & $5.95\,e-3$ & $7.91\,e-4$ & $1.55\,e-2$ & $3.18\,e-2$ \\
3  & $4.24\,e-3$ & $4.01\,e-3$ & $3.28\,e-3$ & $1.77\,e-2$ & $2.74\,e-2$ \\
4  & $2.76\,e-3$ & $5.84\,e-3$ & $3.43\,e-3$ & $1.65\,e-2$ & $3.25\,e-2$ \\
5  & $3.17\,e-3$ & $5.55\,e-3$ & $1.77\,e-3$ & $1.65\,e-2$ & $2.87\,e-2$ \\
\bottomrule
\end{tabular}
\label{tab:grid}
\end{table}

Let observational data be defined on the domain $[-1,1]\times[0,1]$ under uniform discretization with $N_x = 1024$ and $N_t = 1000$. Consider the initial conditions from Fig. \ref{fig:diff_u0}(a) and the reference $\gamma_1\in\left\{0.02, 0.04, 0.6, 0.08, 0.1\right\}$. We evaluate the relative error in $\gamma_1$ under the golden section search method within $\Gamma = [0,0.1]$.
 Table \ref{tab:grid} presents the results, indicating that the inverse problem \eqref{eq:inverse problem} is solved effectively.

\subsection{System for nonconserved parameters}
\label{subsec:noncon}
The DOOL up to here has been established on the system for conserved parameters. 
Now we explain how it can work in parallel for the nonconserved case. 
To do so, we use the following example.

\begin{example}[Allen--Cahn equation]\label{ex:AC}
We consider the Allen--Cahn equation:
\begin{equation}\label{eq:AC}
	\begin{cases}
		\partial_t u(x,t) = \gamma_1  \partial_{xx} u(x,t) - \gamma_2\left(u(x,t)^3 - u(x,t)\right), & x \in (-I, I),\; t \in (0,T], \\
		u(-I,t)=u(I,t),  \ 
		\partial_{x}u(-I,t)=\partial_{x}u(I,t),  \, & t \in [0,T], \\
		u(x,0)=u_0(x), & x\in [-I, I],
	\end{cases}
\end{equation}
which is a popular phase-field model for diffusion-reaction. Here we take $I=1$ and 
$u_0(x)=\tfrac{1}{2}\cos(\pi x)$.
The reference solution is obtained with a spectral solver \cite{ETD}. 
\end{example}

Let us first briefly go through the OVP process for this system \cite{Onsagerchen}. The free energy functional is defined as
\[
\mathcal{E}(u) = \int_{\Omega}
\Bigl(\frac{\gamma_1}{2}(\partial_{x}u)^{2}
+\frac{\gamma_2}{4}\bigl(u^{2}-1\bigr)^{2}\Bigr)\,dx,
\]
and the dissipation function is given by
$
\Phi(\partial_{t}u)=\frac{1}{2}\int_{\Omega} {(\partial_{t}u)}^{2}\,dx.$
The Rayleighian functional is then given by
\begin{align}\label{eq:Rayleighian ac}
	\mathcal{R}(u,\partial_{t}u)  &= \dot{\mathcal{E}}(u,\partial_tu)+\Phi(\partial_{t}u) \nonumber\\
	&=\int_{\Omega}
	\bigl(-\gamma_1 \partial_{xx} u+\gamma_2(u^{3}-u)\bigr)\partial_{t}u\,dx
	+\frac{1}{2}\int_{\Omega} {(\partial_{t}u)}^{2}\,dx.
\end{align}
By the OVP for the nonconserved case, i.e., \eqref{ovp noncon}, the dynamical equation of the system can be derived from
$
\delta \mathcal{R}/\delta (\partial_{t} u)=0.
$

Now to utilize DOOL, we can artificially define a `flux field' $j$ with $\partial_{t} u-j=0$. Then, \eqref{eq:Rayleighian ac} suggests
\begin{align}\label{eq:Rayleighian ac 1}
	\widetilde{\mathcal{R}}(u,j) 
	=\int_{\Omega}
	\bigl(-\gamma_1 \partial_{xx} u+\gamma_2(u^{3}-u)\bigr)j\,dx
	+\frac{1}{2}\int_{\Omega} {j}^{2}\,dx.
\end{align}
This enables Algorithm  \ref{alg:workflow} to work, where
the operator network $\mathcal{G}_\theta$ is still trained to approximate the relation
$\mathcal{G} : u \mapsto j$ by mimizing (\ref{eq:Rayleighian ac 1}). 
Under this configuration, the temporal discretization takes the form: $u^{n+1} = u^{n} + \Delta t \, {j}^{n}$.

\begin{figure}[h!]
	\centerline{
		\psfig{figure=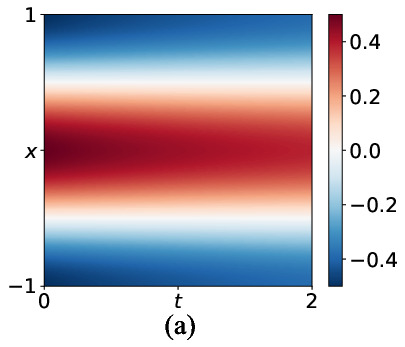,width=1.3in,height=1.1in,angle=0}
		\hspace{-0.1in}
		\psfig{figure=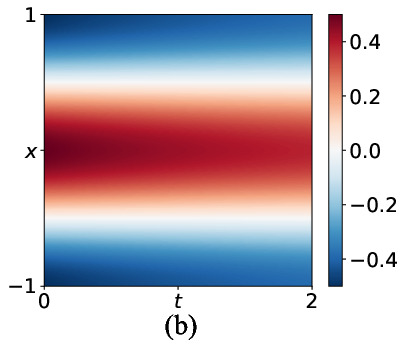,width=1.3in,height=1.1in,angle=0}
		\hspace{-0.1in}
		\psfig{figure=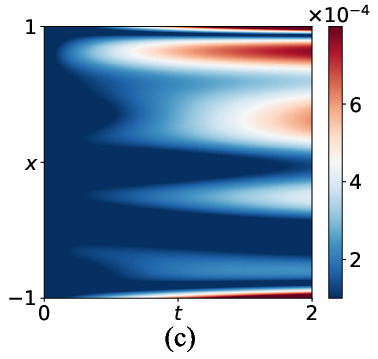,width=1.25in,height=1.1in,angle=0}
            \hspace{-0.1in}
		\psfig{figure=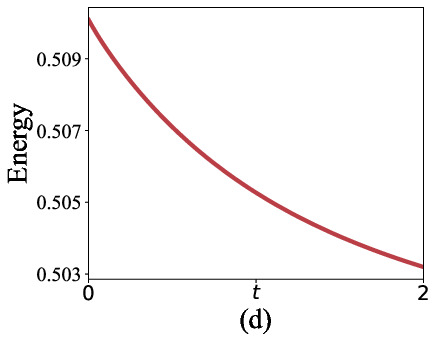,width=1.3in,height=1.1in,angle=0}
	}

    \centerline{
		\psfig{figure=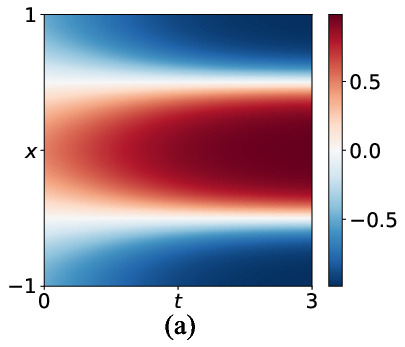,width=1.3in,height=1.1in,angle=0}
		\hspace{-0.1in}
		\psfig{figure=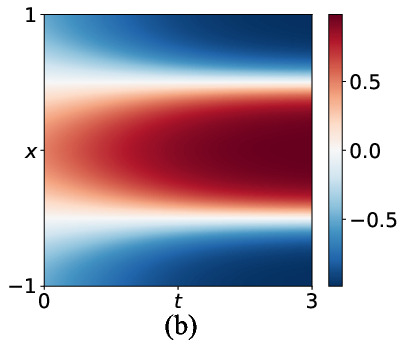,width=1.3in,height=1.1in,angle=0}
		\hspace{-0.1in}
		\psfig{figure=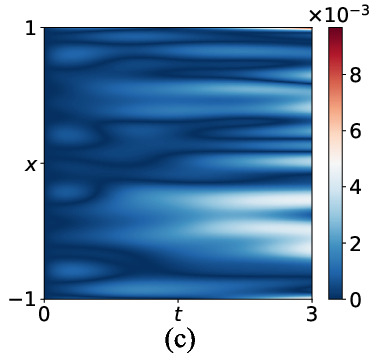,width=1.25in,height=1.1in,angle=0}
          \hspace{-0.1in}
		\psfig{figure=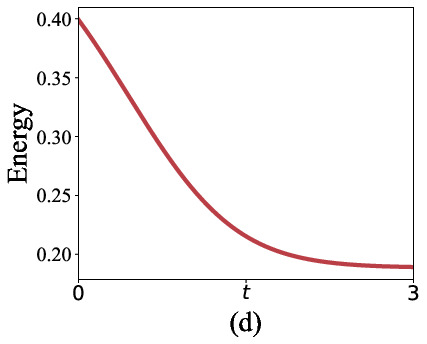,width=1.3in,height=1.1in,angle=0}
	}
	\caption{Results for the Allen--Cahn equation with $\gamma_1 = 0.1$ (1st row) and $\gamma_1 = 0.01$ (2nd row) in Example \ref{ex:AC}: (a) exact solution $u$; (b) numerical solutions $u^n$; (c) pointwise error of $u$; (d) energy decay over time. 
    }\label{fig:Onsager_AC_1}
\end{figure}

Let $\gamma_1 =0.1$, $\gamma_2 =1$ in \eqref{eq:AC}.
We train the operator network by using the Fourier basis with the truncation order $K=9$ and $N_b=100$ training samples.
The hyper-parameters are set as $L = 4$, $W = 70$ and $p=120$. 
After training, problem (\ref{eq:AC}) is solved for $u$ over time to $t=2$ with $\Delta t=0.001$.
We show the exact and numerical solutions obtained by our method in Fig. \ref{fig:Onsager_AC_1}, along with the corresponding pointwise error and the evolution of the free energy over time. 
Additionally, 
the case for $\gamma_1=0.01$ till $t=3$ is also shown in Fig. \ref{fig:Onsager_AC_1}. 

As can be seen from the numerical outcome, the proposed DOOL on the system for nonconserved parameters works as well as in the conserved case. The other results previously established can also be extended to this case. 

\subsection{Second-order wave model with dissipation: deep least-action method} Dissipation also occurs widely in wave phenomena. However, the (second-order in time) wave-type models do not follow the OVP. Its variation comes instead from an earlier classical principle known as the principle of least action \cite{Arnold}. In this circumstance, 
we have recently proposed the deep least action method (DLAM) \cite{DLAM}, which applies to solve the initial-terminal value problem of energy-conservative second-order differential models, including Newtonian dynamics and wave equations. 
DLAM trains the network for predicting the solution by directly minimizing the action functional \cite{DLAM}, while
its extension to dissipative systems remains unexplored. Here, in complement to OVP, we demonstrate how DLAM could be done for wave equations with dissipation. We consider the following illustrative example.

\begin{example}[Damped wave equation]\label{ex:DLAM_PDE}
	 We consider a linear wave equation with  damping 
\begin{equation}\label{eq:DLAM_pde}
	\begin{cases}
		\partial_{tt} u(x,t) + 2\partial_{t}u(x,t) = \partial_{xx}u(x,t), & x\in(-I, I),\; t\in(0,T],\\
		u(x,0)=f(x),\; u(x,T)=g(x),     & x\in(-I, I),\\
		u(-\pi,t)=u(\pi,t),       & t\in[0,T].
	\end{cases}
\end{equation}
Let $I=\pi$, $T=1$, $f(x)=\cos(x)$ and $g(x)=0$,  then the analytical solution of \eqref{ex:DLAM_PDE} is $u(x,t)=(1-{t}/{T})\mathrm{e}^{-t}\cos(x)$. Note that the initial-terminal values are request of the least action principle. In applications, they are considered as observational data.
\end{example}

The principle of least action means that the configuration of $u$ in \eqref{eq:DLAM_pde} with the given initial state $f(x)$ at time $t = 0$ and  terminal state $g(x)$ at $t = T$, must ``minimize" the  \textit{action} functional:
\begin{equation}\label{eq:action}
	\mathcal{S} = \int_{0}^{T}\int_{\Omega} \mathcal{L}[u({x},t), \partial_t{u}({x},t), \partial_x u({x},t)] \, d{x} dt,
\end{equation}
where $\mathcal{L}$ denotes the Lagrangian density function of the system.
The key to adapting DLAM to dissipative wave systems lies in the proper definition of action. Following the work of Bersani et al. \cite{Bersani}, this can be addressed by introducing an exponential damping factor into the Lagrangian of the wave model without damping term. Thus, it yields here $\mathcal{L} = \mathrm{e}^{2t}\left((\partial_tu)^{2} - (\partial_xu)^{2}\right)$.
The variational calculus from
$\frac{\delta\mathcal{S}}{\delta u}=0 $
gives the dynamical equation in \eqref{eq:DLAM_pde}.

For the initial-terminal value problem, DLAM proposes a normalized deep neural network (NDNN) to approximate $u$, specifically by adding an additional transformation/normalization layer to the classical fully connected neural network to enforce hard constraints on the initial and terminal values.
The architecture of NDNN reads
\begin{equation}\label{NDNN}
	u_\theta(\by_0) = \Phi \circ {\bf F}_{L+1} \circ \sigma \circ {\bf F}_{L} \circ \sigma \circ\cdots \circ {\bf F}_{2} \circ \sigma \circ {\bf F}_{1}(\by_0), \quad \theta:=\left\{W_l,b_l\right\},
\end{equation}
where $\sigma$ is the activation function, ${\bf F}_l(\by_{l-1}) = W_l \by_{l-1} + b_l$ is the affine transformation with $\by_{l-1} \in \mathbb{R}^{n_{l-1}}$, $1\leq l \leq L+1$, and $\Phi$ is the normalization layer
	$\Phi(\mathbf{y}) := \sin({t\pi}/{T})\mathbf{y} + \frac{\sin(T-t)}{\sin(T)}f(x) +\frac{\sin(t)}{\sin(T)}g(x).$
By the least action principle, the action \eqref{eq:action} that $u$ needs to minimize can naturally serve as the loss function.
We select a fixed uniform grid of $N_x \times N_t$ points $\{x_l,t_k\}$ with spatial discretization $x_l =-I+  2Il/N_x,\  l=1, \dots, N_x$ and temporal discretization  $t_k =  kT/N_t,\  k=1, \dots, N_t$, where $N_x = N_t = 128$. 
Based on this discrete point set, the practical loss function can be defined as
\begin{equation*}
	\text{Loss}(\theta) = \frac{2I\times T}{N_x\times N_t} \sum_{l=1}^{N_x}\sum_{k=1}^{N_t} \mathrm{e}^{2t_k} \left( \left(\partial_t u_\theta(x_l,t_k)\right)^{2} - \left(\partial_x u_\theta(x_l,t_k)\right)^{2} \right).
\end{equation*}

\begin{figure}[h!]
	\centerline{
		\psfig{figure=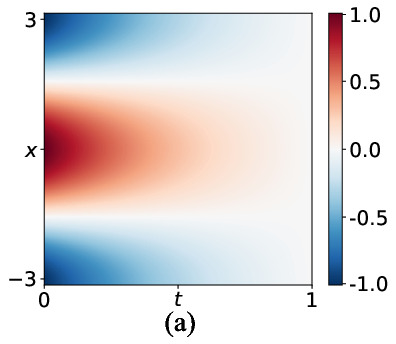,width=1.3in,height=1.1in,angle=0}
		\hspace{-0.1in}
		\psfig{figure=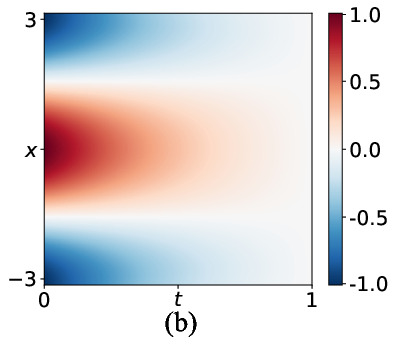,width=1.3in,height=1.1in,angle=0}
		\hspace{-0.1in}
		\psfig{figure=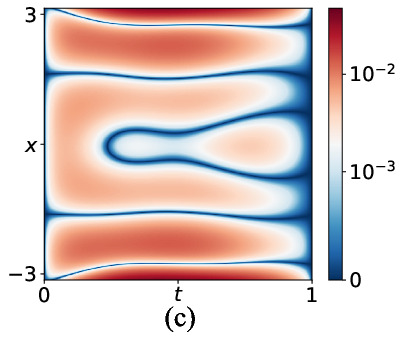,width=1.3in,height=1.1in,angle=0}
		\hspace{-0.1in}
		\psfig{figure=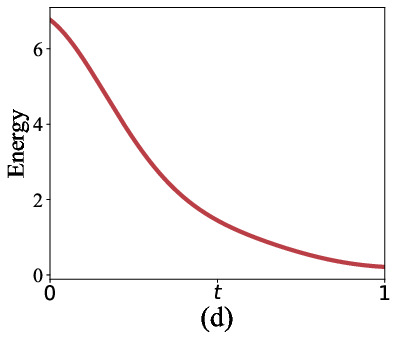,width=1.3in,height=1.1in,angle=0}
	}
	\caption{Results for Example \ref{ex:DLAM_PDE} with $L=4$, $W=70$: (a) exact solution; (b) numerical solution; (c) pointwise error; (d) energy decay over time. }\label{fig:DLAM_PDE}
\end{figure}

Set the activate function as $\sigma=\tanh$ and take the hyper-parameters as $L=4$, $W=70$.
We show the exact and numerical solutions obtained by DLAM in Fig. \ref{fig:DLAM_PDE}, along with the pointwise error and the energy behavior in time. 
These results demonstrate that DLAM performs effectively in solving the dissipative wave equation \eqref{ex:DLAM_PDE}, accurately recovering $u(x,t)$ with clear energy dissipation.

Subsequent operator learning tasks within DLAM can be carried out similarly to DOOL, though the details are omitted here for brevity. Together, DLAM and DOOL cover the majority of dissipative problems encountered in applications, offering effective prediction of solutions through efficient unsupervised learning.


\section{Conclusion}
\label{sec:conclusions}
In this study, we considered deep operating learn towards general dissipative models. Based on the Onsager variational principle (OVP), a novel unsupervised framework for operator learning is proposed, called deep Onsager operator learning (DOOL).
The architecture of DeepONet is adopted to approximate the targeted operator. By minimizing the OVP-defined Rayleighian functional, the operator network is trained in a completely unsupervised manner to learn the inherent constitutive relation between the state and the flux of the concerned dissipative system. This does not require any labeled data from prior simulations. The trained operator network is then utilized in discretized conservation/change laws for an external and explicit time stepping, which naturally provides temporal extrapolation capability. 
Numerical experiments on serval typical dissipative equations (Fokker-Planck, Allen-Cahn, Cahn-Hilliard etc.) were conducted to demonstrate the effectiveness of DOOL.
Comparisons were made with the supervising-mannered DeepONet and MIONet to show the improved accuracy and efficiency of DOOL. In addition, monotone energy dissipation is numerically observed and analyzed. 
Applications include a physical parameter inversion task, and 
extensions include a deep least-action method for the second-order wave models with dissipation which do not directly follow OVP.



\appendix
\section{Analysis of energy decrease}
This appendix provides some analysis for the observed energy dissipation under DOOL. We note that the free energy functional reads as $\mathcal{E} (u):=\int_{\Omega} \xi(u)\,d\bx$ and $\Phi(u,\,\boldsymbol{j})$ is nonnegative satisfying $\Phi(u,\,\boldsymbol{0})=0$. Boundary conditions are assumed to eliminate flux contributions.

\subsection{System for conserved parameters}
With conserved equation $\partial_t u + \nabla \cdot \boldsymbol{j} = 0$, the semi-discrete scheme minimizes:
\begin{equation*}
    \boldsymbol{j}^n = \underset{\boldsymbol{j}}{\operatorname{argmin}} \int_{\Omega} \nabla \xi^{\prime}(u^n)  \cdot \boldsymbol{j}  \, d\bx + \Phi(u^n,\, \boldsymbol{j}),  \qquad 
		\frac{u^{n+1} - u^{n}}{\Delta t} + \nabla \cdot \boldsymbol{j}^{n} = 0.
\end{equation*}
From the minimization, we can see that 
\begin{equation*}
     \int_{\Omega}  \nabla\xi^{\prime}(u^n) \cdot
 \boldsymbol{j}^n  \, d\bx + \Phi(u^n,\, \boldsymbol{j}^n) 
     \leq
     \int_{\Omega} \nabla\xi^{\prime}(u^n) \cdot \boldsymbol{0}  \, d\bx + \Phi(u^n,\, \boldsymbol{0}) = 0,   
\end{equation*}
which implies:
$     \int_{\Omega}  \nabla \xi^{\prime}(u^n) \cdot
 \boldsymbol{j}^n  \, d\bx 
     \leq -\Phi(u^n,\, \boldsymbol{j}^n)   \leq  0.
$
Integration-by-parts gives
\begin{align*}
\int_{\Omega}  \nabla \xi^{\prime}(u^n) \cdot
 \boldsymbol{j}^n  \, d\bx &=
\int_{\Omega} \xi^{\prime}(u^n) \cdot
 (-\nabla \cdot \boldsymbol{j}^n)  \, d\bx =
\int_{\Omega} \xi^{\prime}(u^n) \frac{u^{n+1} - u^{n}}{\Delta t}  \, d\bx
\leq  0.
\end{align*}
Then, we find for the numerical energy that
\[
\mathcal{E} (u^{n+1})-\mathcal{E} (u^{n}) = \int_{\Omega} \left[ \xi(u^{n+1}) - \xi(u^n) \right]  d\bx \approx \int_{\Omega} \xi^{\prime}(u^n) (u^{n+1} - u^{n})  \, d\bx \leq 0,
\]
which provides a preliminary indication of the energy dissipation characteristics of the DOOL method. Rigorous proof will be a future task.

\subsection{System for nonconserved parameters}
With
\begin{equation*}
    \boldsymbol{j}^n = \underset{\boldsymbol{j}}{\operatorname{argmin}} \int_{\Omega}  \xi^{\prime}(u^n)  \cdot \boldsymbol{j}  \, d\bx + \Phi(u^n,\, \boldsymbol{j}),  \qquad 
		\frac{u^{n+1} - u^{n}}{\Delta t} =  \boldsymbol{j}^{n},
\end{equation*}
we can find  
\begin{equation*}
     \int_{\Omega}  \xi^{\prime}(u^n) \cdot
 \boldsymbol{j}^n  \, d\bx \leq \int_{\Omega}  \xi^{\prime}(u^n) \cdot
 \boldsymbol{j}^n  \, d\bx + \Phi(u^n,\, \boldsymbol{j}^n) 
     \leq
     \int_{\Omega} \xi^{\prime}(u^n) \cdot \boldsymbol{0}  \, d\bx + \Phi(u^n,\, \boldsymbol{0}) = 0,  
\end{equation*}
which implies 
$
\int_{\Omega} \xi^{\prime}(u^n) \left(u^{n+1} - u^{n}\right)  \, d\bx
\leq  0.$

\bibliographystyle{siamplain}
\bibliographystyle{model1-num-names}

\end{sloppypar}
\end{document}

%% file: ex_shared.tex
\usepackage{lipsum}
\usepackage{amsfonts}
\usepackage{graphicx}
\usepackage{epstopdf}
\usepackage{algorithmic}
\ifpdf
  \DeclareGraphicsExtensions{.eps,.pdf,.png,.jpg}
\else
  \DeclareGraphicsExtensions{.eps}
\fi

\newsiamremark{remark}{Remark}
\newsiamremark{hypothesis}{Hypothesis}
\crefname{hypothesis}{Hypothesis}{Hypotheses}
\newsiamthm{claim}{Claim}

\headers{Onsager  principle for unsupervised operator learning}{Zhipeng Chang, Zhenye Wen, Xiaofei Zhao}

\title{Unsupervised operator learning approach for dissipative equations via Onsager principle 
\thanks{Submitted to the editors DATE. \funding{This work is supported by National
Key Research and
Development Program
of China
(No. 2024YFE03240400), National MCF Energy R\&D Program, National Natural Science Foundation of China
(Nos. 42450275, 12271413)}
}}

\author{Zhipeng Chang\thanks{School of Mathematics and Statistics, Wuhan University, 430072 Wuhan, China (\email{changzhipeng@whu.edu.cn}).}
\and Zhenye Wen \thanks{School of Mathematics and Statistics, Wuhan University, 430072 Wuhan, China (\email{wenzhy@whu.edu.cn}).}
  \and Xiaofei Zhao\thanks{School of Mathematics and Statistics, and Hubei Key Laboratory of Computational Science, Wuhan University, Wuhan, China (\email{matzhxf@whu.edu.cn}).}}

\usepackage{amsopn}
